\pdfoutput=1
\documentclass[10pt,logo,copyright]{nvidiatechreport}
\usepackage[authoryear,sort&compress,round]{natbib}
\usepackage{parskip}   

\usepackage{amsmath,amsfonts,bm}

\def\1{\bm{1}}

\DeclareMathAlphabet{\mathsfit}{\encodingdefault}{\sfdefault}{m}{sl}
\SetMathAlphabet{\mathsfit}{bold}{\encodingdefault}{\sfdefault}{bx}{n}

\usepackage{mathtools}
\usepackage{multirow}
\usepackage{tabularx}
\usepackage{array}
\usepackage{colortbl}
\usepackage{etoolbox}  
\usepackage{pifont}  
\usepackage{float}
\usepackage{tikz}
\usetikzlibrary{arrows.meta,positioning,fit,calc,backgrounds}

\hypersetup{pdftitle={MAWAM: A Multi-Agent World-Action Model for Coordinated Manipulation}}

\makeatletter
\providecommand{\@LN}[2]{}
\providecommand{\@LN@col}[1]{}
\makeatother

\newcommand{\method}{\textsc{Mawam}\xspace}

\newcommand{\robofactory}{\textsc{RoboFactory}\xspace}
\newcommand{\warmup}{Warm-up\xspace}
\newcommand{\coord}{Team training\xspace}
\newcommand{\coordstage}{Team training\xspace}

\newcommand{\sym}{\textsc{Sym}}
\newcommand{\asym}{\textsc{Asym}}
\newcommand{\mixed}{\textsc{Mixed}}
\newcommand{\best}[1]{\textbf{#1}}
\newcommand{\second}[1]{\underline{#1}}
\definecolor{ourshade}{HTML}{E4F1CC}  
\definecolor{nvgreen}{RGB}{118,185,0}  
\definecolor{ablchg}{HTML}{3A7D00}  
\newcommand{\abon}{\ding{51}}

\newcolumntype{Y}{>{\centering\arraybackslash}X}
\newif\ifsetable
\makeatletter
\newcommand{\sefont}{\ifdim\f@size pt>8.5pt \fontsize{7}{8}\selectfont\else\ifdim\f@size pt>6.5pt \fontsize{6}{7}\selectfont\else\fontsize{5}{6}\selectfont\fi\fi}  
\makeatother
\newcommand{\sepm}[1]{\ifsetable{\sefont\,$\pm$\,#1}\else~$\pm$~#1\fi}
\AtBeginEnvironment{table}{\setabletrue}
\AtBeginEnvironment{table*}{\setabletrue}
\newcommand{\sepmtablesonly}{\renewcommand{\sepm}[1]{\ifsetable{\sefont\,$\pm$\,##1}\fi}}

\newcommand{\SimMawamLiftPct}{100.0\sepm{0.0}}  

\newcommand{\SimMawamPassShoePct}{100.0\sepm{0.0}}  

\newcommand{\SimMawamPlaceFoodPct}{78.0\sepm{4.1}}  

\newcommand{\SimMawamTwoStackPct}{93.0\sepm{2.6}}  

\newcommand{\SimMawamCameraAlignPct}{100.0\sepm{0.0}}  

\newcommand{\SimMawamThreeStackPct}{59.0\sepm{4.9}}  

\newcommand{\SimMawamTakePhotoPct}{97.0\sepm{1.7}}  

\newcommand{\SimMawamPipelinePct}{95.0\sepm{2.2}}  

\newcommand{\SimMawamMacro}{90.3\sepm{0.9}}  

\newcommand{\MAVLALift}{98.0}
\newcommand{\MAVLAPassShoe}{91.0}
\newcommand{\MAVLAPlaceFood}{59.0}
\newcommand{\MAVLATwoStack}{86.0}
\newcommand{\MAVLACameraAlign}{96.0}
\newcommand{\MAVLAThreeStack}{58.0}
\newcommand{\MAVLATakePhoto}{82.0}
\newcommand{\MAVLAMacro}{83.4}
\newcommand{\MAVLAPipeline}{97.0}

\newcommand{\GauDPLift}{72.0}
\newcommand{\GauDPPassShoe}{n/a}
\newcommand{\GauDPPlaceFood}{15.0}
\newcommand{\GauDPTwoStack}{2.0}
\newcommand{\GauDPCameraAlign}{26.0}
\newcommand{\GauDPThreeStack}{0.0}
\newcommand{\GauDPTakePhoto}{3.0}
\newcommand{\GauDPPipeline}{n/a}
\newcommand{\GauDPMacro}{n/a}

\newcommand{\LTLDiffLift}{96.0}
\newcommand{\LTLDiffPassShoe}{7.0}
\newcommand{\LTLDiffPlaceFood}{25.0}
\newcommand{\LTLDiffTwoStack}{60.0}
\newcommand{\LTLDiffCameraAlign}{67.0}
\newcommand{\LTLDiffThreeStack}{7.0}
\newcommand{\LTLDiffTakePhoto}{24.0}
\newcommand{\LTLDiffPipeline}{0.0}
\newcommand{\LTLDiffMacro}{35.8}
\newcommand{\CLSDPLift}{61.0}
\newcommand{\CLSDPPassShoe}{n/a}
\newcommand{\CLSDPPlaceFood}{43.0}
\newcommand{\CLSDPTwoStack}{39.0}
\newcommand{\CLSDPCameraAlign}{55.0}
\newcommand{\CLSDPThreeStack}{20.0}
\newcommand{\CLSDPTakePhoto}{8.0}
\newcommand{\CLSDPPipeline}{n/a}
\newcommand{\CLSDPMacro}{n/a}

\newcommand{\SimIndependentLift}{99.0\sepm{1.0}}  
\newcommand{\SimIndependentPlaceFood}{54.0\sepm{5.0}}  
\newcommand{\SimIndependentCameraAlign}{99.0\sepm{1.0}}  
\newcommand{\SimIndependentPassShoe}{83.0\sepm{3.8}}  
\newcommand{\SimIndependentPipeline}{39.0\sepm{4.9}}  

\newcommand{\SimCentralPassShoePct}{79.0\sepm{4.1}}  

\newcommand{\SimCentralPlaceFoodPct}{54.0\sepm{5.0}}  

\newcommand{\SimCentralPipelinePct}{0.0\sepm{0.0}}  
\newcommand{\SimCentralPipelineSixK}{0/100}  

\newcommand{\SimPerAgentActionLift}{99.0\sepm{1.0}}  
\newcommand{\SimPerAgentActionPlaceFood}{47.0\sepm{5.0}}  
\newcommand{\SimPerAgentActionCameraAlign}{94.0\sepm{2.4}}  
\newcommand{\SimPerAgentActionTwoStack}{92.0\sepm{2.7}}  
\newcommand{\SimPerAgentActionPipeline}{91.0\sepm{2.9}}  

\newcommand{\SimNoWarmupLift}{100.0\sepm{0.0}}  
\newcommand{\SimNoWarmupPlaceFood}{45.0\sepm{5.0}}  
\newcommand{\SimNoWarmupCameraAlign}{92.0\sepm{2.7}}  
\newcommand{\SimNoWarmupPassShoe}{98.0\sepm{1.4}}  
\newcommand{\SimNoWarmupPipeline}{34.0\sepm{4.7}}  

\newcommand{\SimNoGlobalLift}{96.0\sepm{2.0}}  
\newcommand{\SimNoGlobalPlaceFood}{62.0\sepm{4.9}}  
\newcommand{\SimNoGlobalCameraAlign}{96.0\sepm{2.0}}  
\newcommand{\SimNoGlobalTwoStack}{95.0\sepm{2.2}}  

\newcommand{\SimFixedSimplexLift}{84.0\sepm{3.7}}  
\newcommand{\SimFixedSimplexPlaceFood}{43.0\sepm{5.0}}  
\newcommand{\SimFixedSimplexCameraAlign}{99.0\sepm{1.0}}  

\newcommand{\SimAbsSlotsLift}{91.0\sepm{2.9}}  
\newcommand{\SimAbsSlotsPlaceFood}{22.0\sepm{4.1}}  
\newcommand{\SimAbsSlotsCameraAlign}{99.0\sepm{1.0}}  

\newcommand{\SimAbsSlotsTakePhoto}{96.0\sepm{2.0}}  
\newcommand{\SimFixedSimplexTakePhoto}{95.0\sepm{2.2}}  
\newcommand{\SimNoRolesPlaceFood}{0.0\sepm{0.0}}  
\newcommand{\SimNoRolesCameraAlign}{8.0\sepm{2.7}}  
\newcommand{\SimRoleSwapPlaceFood}{0.0\sepm{0.0}}  
\newcommand{\SimRoleSwapCameraAlign}{0.0\sepm{0.0}}  
\newcommand{\SimSharedHolderCameraAlign}{100.0\sepm{0.0}}  

\newcommand{\RealMawamUmbrella}{86.7\sepm{6.2}}
\newcommand{\RealMawamTowel}{63.3\sepm{8.8}}
\newcommand{\RealMawamLaundry}{53.3\sepm{9.1}}
\newcommand{\RealMawamMacro}{67.8\sepm{4.7}}

\newcommand{\RealGauDPUmbrella}{50.0\sepm{9.1}}
\newcommand{\RealGauDPTowel}{23.3\sepm{7.7}}
\newcommand{\RealGauDPLaundry}{13.3\sepm{6.2}}
\newcommand{\RealGauDPMacro}{28.9\sepm{4.5}}

\newcommand{\RealMAVLAUmbrella}{70.0\sepm{8.4}}
\newcommand{\RealMAVLATowel}{46.7\sepm{9.1}}
\newcommand{\RealMAVLALaundry}{30.0\sepm{8.4}}
\newcommand{\RealMAVLAMacro}{48.9\sepm{5.0}}
\newcommand{\RealIndepLaundry}{36.7\sepm{8.8}}
\newcommand{\RealIndepTowel}{50.0\sepm{9.1}}
\newcommand{\RealIndepUmbrella}{70.0\sepm{8.4}}
\newcommand{\RealIndepMacro}{52.2\sepm{5.1}}

\newcommand{\SimTTNoFutLift}{100.0\sepm{0.0}}  
\newcommand{\SimTTNoFutPassShoe}{1.0\sepm{1.0}}  
\newcommand{\SimTTNoFutPlaceFood}{--}  
\newcommand{\SimTTNoFutTwoStack}{0.0\sepm{0.0}}  
\newcommand{\SimTTNoFutCameraAlign}{98.0\sepm{1.4}}  
\newcommand{\SimTTNoFutThreeStack}{20.0\sepm{4.0}}  
\newcommand{\SimTTNoFutTakePhoto}{90.0\sepm{3.0}}  
\newcommand{\SimTTNoFutPipeline}{0.0\sepm{0.0}}  
\newcommand{\SimTTFrozenLift}{100.0\sepm{0.0}}  
\newcommand{\SimTTFrozenPassShoe}{100.0\sepm{0.0}}  
\newcommand{\SimTTFrozenPlaceFood}{70.0\sepm{4.6}}  
\newcommand{\SimTTFrozenTwoStack}{98.0\sepm{1.4}}  
\newcommand{\SimTTFrozenCameraAlign}{100.0\sepm{0.0}}  
\newcommand{\SimTTFrozenThreeStack}{54.0\sepm{5.0}}  
\newcommand{\SimTTFrozenTakePhoto}{98.0\sepm{1.4}}  
\newcommand{\SimTTFrozenPipeline}{98.0\sepm{1.4}}  

\AtBeginEnvironment{table}{\arrayrulecolor{nvidiagreen}}
\AtBeginEnvironment{table*}{\arrayrulecolor{nvidiagreen}}

\definecolor{revisionred}{RGB}{170,25,35}
\newcommand{\rev}[1]{\ifdefined\showrevisions\textcolor{revisionred}{#1}\else#1\fi}

\titleformat{\paragraph}[runin]{\normalfont\normalsize\bfseries}{\theparagraph}{1em}{#1}
\titlespacing*{\paragraph}{0pt}{1.5ex plus 1ex minus .2ex}{0.5em}

\titleformat{name=\section,numberless}{\Large\bfseries\headingfont}{}{0pt}{#1}[]

\newlength{\nvcolfloatwidth}
\makeatletter
\AtBeginDocument{\g@addto@macro\@floatboxreset{\setlength{\columnwidth}{\nvcolfloatwidth}}}
\makeatother

\makeatletter
\def\nv@stripbang#1{\nv@sb#1!\@nil}
\def\nv@sb#1!#2\@nil{#1}
\renewenvironment{figure*}[1][tbp]{\edef\nv@tmp{\noexpand\@float{figure}[\nv@stripbang{#1}]}\nv@tmp}{\end@float}
\renewenvironment{table*}[1][tbp]{\edef\nv@tmp{\noexpand\@float{table}[\nv@stripbang{#1}]}\nv@tmp}{\end@float}
\makeatother

\makeatletter
\renewcommand{\abscontent}{%
  \Needspace*{4\baselineskip}%
  \noindent{\large\bfseries\headingfont Abstract}\par\nopagebreak
  {\absfont\clubpenalty=10000 \widowpenalty=10000 \noindent\theabstract\par}
  \@ifundefined{@keywords}{}{\vskip1em \noindent \keywordsfont  Keywords: \@keywords}}
\makeatother

\title{MAWAM: A Multi-Agent World-Action Model for Coordinated Manipulation}

\author{
  Xianzhe~Fan$^{1,2}$
  Chi-Pin~Huang$^{1}$
  Fu-En~Yang$^{1}$
  Jing Wang$^{1}$
  Yecheng~Wu$^{1,3}$
  Yu-Cheng~Chou$^{1}$
  Johan~Bjorck$^{1}$
   Linxi~``Jim''~Fan$^{1}$
   Jan~Kautz$^{1}$
   Yu-Chiang~Frank~Wang$^{1}$
   Zhiding~Yu$^{1}$
   Hengshuang~Zhao$^{2}$
   Zhiqi~Li$^{1}$ \\
  \small $^{1}$NVIDIA \quad $^{2}$ The University of Hong Kong \quad $^{3}$ MIT \\
  {\small \href{https://research.nvidia.com/labs/lpr/mawam/}{\texttt{https://research.nvidia.com/labs/lpr/mawam}}}
}

\begin{abstract}
Every move in a robot team changes the world in which its teammates must act. Coordinated manipulation therefore calls for predicting the team's actions together with the physical future they jointly create. We introduce \method, a multi-agent world-action model that jointly generates each robot's future views and executable action chunks in a single sample. Built from a pretrained single-robot world-action backbone, \method jointly denoises all robots' action and future-view tokens through cross-agent attention, while preserving a separate control interface for each robot. \method achieves the highest success on all three real-robot tasks (laundry-bag collection, bath-towel folding, and umbrella handover), with 67.8\% mean success, 18.9 percentage points above MA-VLA trained on the same demonstrations and cameras. In simulation, \method reaches 90.3\% mean success on eight tasks with teams of up to four arms. Ablations show that cross-agent attention and team-stage future-video supervision both improve coordinated control: without attention, a four-arm relay falls from 95.0\% to 39.0\%, and without this supervision, Place Food falls from 78.0\% to 47.0\%.
\end{abstract}

\begin{document}

\maketitle

\begin{center}
\captionsetup{type=figure}
\includegraphics[width=\textwidth]{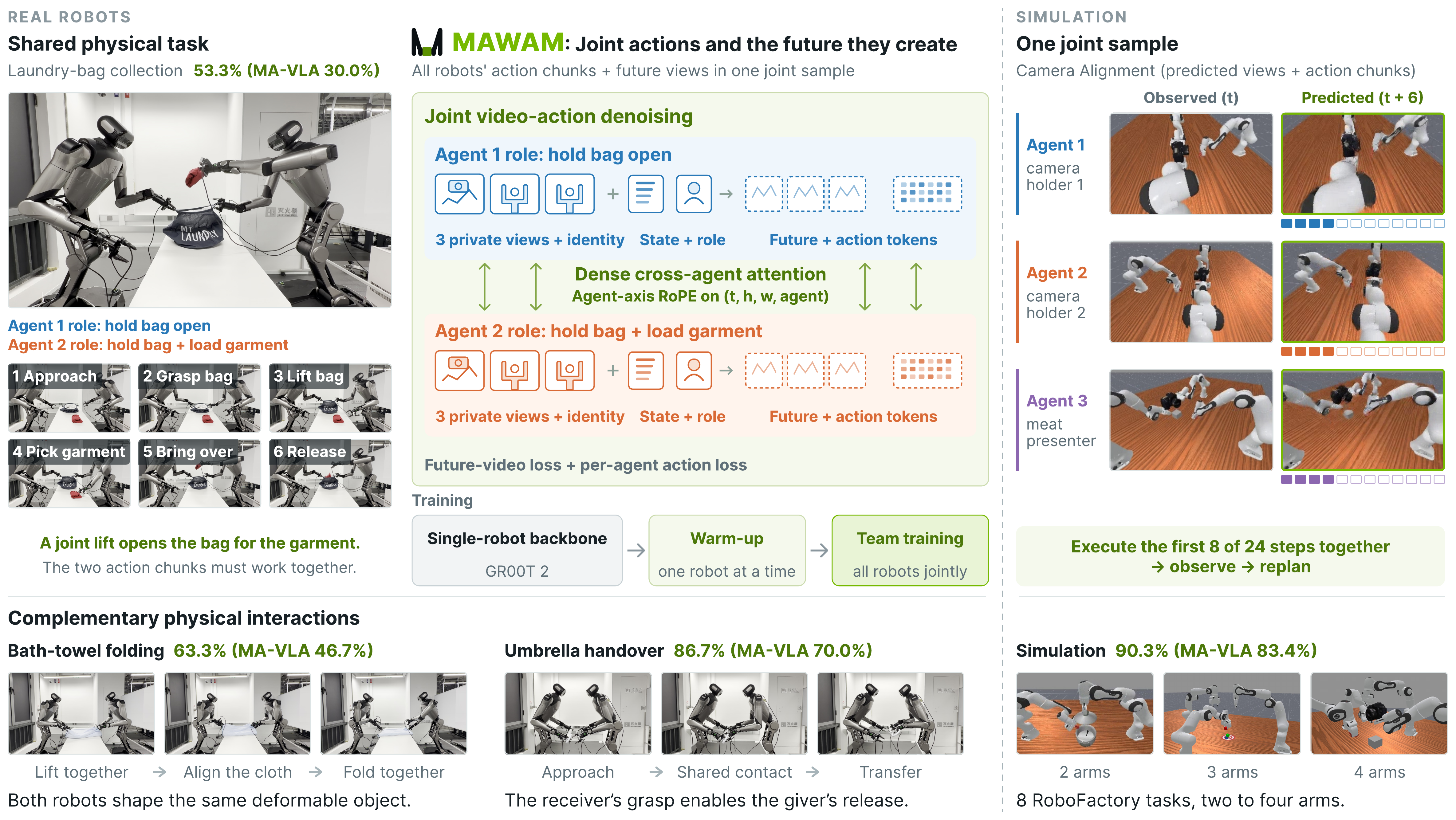}
\captionof{figure}{\textbf{One sample, the whole team's future.} \method generates every robot's future camera views and action chunk together; training it to predict the whole team's future teaches each robot to coordinate with its teammates. \emph{Left:} two real robots, each with three private cameras and no shared global camera; in laundry-bag collection both lift the bag to hold it open while Agent~2 loads the garment. \emph{Right:} two to four simulated arms; in Camera Alignment each agent's private view is predicted six control steps ahead within a 24-step joint sample, whose first eight steps are executed before replanning (a read-only shared global camera provides context).}
\label{fig:teaser}
\end{center}
\vspace{0.1in}

\abscontent\unskip\unskip  

\section{Introduction}
\label{sec:introduction}

Robots will increasingly work in teams to tackle complex tasks in the physical world. In homes and factories, robots may need to hand objects to one another, jointly manipulate deformable materials, or carry loads together. In these tasks, each robot observes the scene from its own viewpoint, but its actions change the shared environment and shape what its teammates can do next. A grasp may alter the shape of an object held by another robot, while a handover may enable a teammate to begin its next operation \citep{mandi2024roco,qin2025robofactory}. Successful collaboration therefore requires robots to coordinate their actions and timing to achieve a shared goal. Viewed this way, collaborative manipulation is a process in which robots jointly shape the future of their shared physical world. This motivates us to formulate it as a multi-agent world-modeling problem: jointly predicting the team's shared future and the action sequences that produce it.

Existing approaches to multi-robot collaboration largely fall into three categories. (1) \textit{Decentralized policy coordination} retains a separate execution policy for each robot while learning coordination mechanisms. Although readily scalable, these policies may lead to idle waiting or oscillatory behavior in tightly coupled tasks because of mismatched coordination patterns or desynchronized execution \citep{pmlr-v305-he25a,dong2026mimic,park2026distilling}. (2) \textit{Joint trajectory and action generation} methods model multi-agent trajectories or generate coordinated controls but lack explicit predictions of their visual consequences \citep{NEURIPS2025_0895de15,zhu2024madiff}. (3) \textit{Language- and logic-guided task coordination} uses planning, dialogue, or temporal logic to organize collaboration, but does not explicitly predict the visual consequences of joint actions \citep{chen2024scalable,kannan2024smart,mandi2024roco,meng2026ltldiff,zhang2026ma}. These methods either decentralize decision-making, lack physical foresight, or rely on task-level guidance; none jointly generates executable team actions and their corresponding visual futures.

World models offer a natural path toward addressing this lack of physical foresight, yet existing research still separates multi-agent future prediction from executable joint control. Multi-agent world models can reconstruct shared states or disentangle per-agent dynamics for planning \citep{zhang2025combo,zhao2026empowering}, and in \textit{video generation} they can produce multi-view-consistent futures \citep{savva2026solaris,zhu2026shareverse,wu2026multiworld,liu2026gamma}. However, these video models treat actions only as conditioning inputs and do not produce policies that can be directly deployed in the physical world \citep{savva2026solaris,wu2026multiworld,liu2026gamma}. In \textit{robotics}, recent world-action models jointly learn visual prediction and action generation, demonstrating that future-prediction representations can improve control by capturing the dependencies between actions and anticipated scene evolution \citep{guo2024prediction,li2025unifiedvideoactionmodel,NEURIPS2025_89a3b655,ye2026world}. However, these models are still organized around a single robot's control stream. Multi-agent video generation models thus predict without producing executable joint control, whereas multi-robot policies act without jointly modeling the visual consequences that each agent will encounter.

This leads to a central question: how can a unified generative model jointly predict a robot team's future observations and generate coordinated actions for manipulation in a shared physical world? We introduce \textbf{\method}, a \textbf{multi-agent world-action model} that learns coordinated control by jointly modeling each robot's actions and the team's future camera views. A shared Transformer connects all agents' action and future-view tokens through cross-agent attention and jointly denoises them. Each generated sample pairs executable action chunks with synchronized future views, predicting both how the team will act and how the scene will evolve (Figure~\ref{fig:teaser}).

Joint prediction needs to preserve which observations and controls belong to each robot while distinguishing its identity from its task responsibility. \method uses an explicit view--agent map to associate multiple private camera streams with their owner, together with per-agent state and action tokens that retain each robot's control interface. Regular-simplex rotary identities assign the same identity code to all views of a robot, an action-to-video bias ties its action tokens to those views, and separate task-role embeddings specify its responsibility. This allows multiple robots to share a task role without merging their observations or controls: for example, two camera holders can share a role while retaining separate states and action outputs.

To adapt GR00T~2~\citep{nvidia2026gr00t2}, a single-robot world-action model, to team control, we use two training stages. During \textit{warm-up}, the model learns from one robot's views, state, and actions at a time, adapting to the cameras and controllers. \textit{Team training} then enables cross-agent attention and role conditioning, using synchronized demonstrations to jointly supervise the robots' actions and future views. Randomized identity assignments and complete-agent-group shuffling expose each role to varied identity codes and sequence positions while preserving each robot's observation--action correspondence. At deployment, joint denoising produces the team's future views and action chunks. The robots execute their actions in synchrony, then use updated observations to condition the next joint prediction.

We evaluate \method on three real-robot tasks: laundry-bag collection, bath-towel folding, and umbrella handover. Using three private cameras per robot and no shared global camera, \method achieves 67.8\% mean success, outperforming MA-VLA by 18.9 percentage points under the same demonstration and camera setup. We further evaluate on eight \robofactory tasks with two to four arms, including camera alignment, collaborative photography, and multi-stage delivery, achieving 90.3\% average success.
Ablations show that removing cross-agent attention reduces four-arm relay success from 95.0\% to 39.0\%, while removing future-video supervision in team training lowers Place Food success from 78.0\% to 47.0\%. Additional ablations demonstrate the benefits of task-role conditioning, randomized agent identities, and single-robot warm-up for coordinated manipulation.

Our main contributions are:
\begin{itemize}
\item We propose \method, a \textit{multi-agent world-action model} that jointly denoises all agents' future-view and action tokens through cross-agent attention. Joint video-action flow matching enables a single sample to contain synchronized visual futures and executable actions for the entire team.
\item We design an \textit{agent-attributed representation} that explicitly associates each robot's private views, state, and actions, combining regular-simplex agent identities with separate task-role embeddings. This preserves each robot's observation and control interfaces while allowing multiple robots to share a task responsibility. We further develop a two-stage training scheme: \textit{single-robot warm-up} adapts the pretrained world-action backbone to the target cameras and controllers, and \textit{synchronized team training} learns cross-robot interactions with randomized identity assignments and agent-group shuffling.
\item We demonstrate coordinated manipulation on three real-robot tasks and eight simulation tasks with two to four arms. \method improves mean success over MA-VLA by 18.9 percentage points on real robots and exceeds its published simulation average by 6.9 percentage points. Ablations examine the representation, future supervision, cross-agent interaction, and training recipe.
\end{itemize}

\section{Related Work} \label{sec:related}

\paragraph{Coordinating robot actions.}
\rev{Existing approaches can be organized around three broad coordination paradigms.
\textit{(1) Decentralized policy coordination.}
These methods retain a separate execution policy for each robot and learn mechanisms for coordinating their behavior. LatentToM aligns consensus representations across agents, while MIMIC-D jointly trains decentralized diffusion policies using local information \citep{pmlr-v305-he25a,dong2026mimic}. CLS-DP distills collaborative dynamics into latent representations that condition each agent's policy \citep{park2026distilling}.
\textit{(2) Joint trajectory and action generation.}
MADiff models joint multi-agent trajectories with diffusion, supporting both centralized control and decentralized execution \citep{zhu2024madiff}. GauDP integrates agents' observations through a shared Gaussian scene representation for coordinated control \citep{NEURIPS2025_0895de15}.
\textit{(3) Language- and logic-guided task coordination.}
CMAS and SMART-LLM use LLMs for action assignment and task planning, while RoCo coordinates robots through dialogue \citep{chen2024scalable,kannan2024smart,mandi2024roco}. LTLDiff uses language-derived temporal logic to guide demonstration generation and multi-agent diffusion policies \citep{meng2026ltldiff}. MA-VLA uses a planner to produce per-arm atomic prompts and a shared VLA executor to generate all arms' actions \citep{zhang2026ma}.
Benchmarks such as \robofactory support systematic evaluation of collaborative manipulation \citep{qin2025robofactory}.}
These approaches coordinate robot actions or task plans, but do not jointly generate executable team actions and their corresponding visual futures. \method unifies the two in a single joint sample, coupling all agents' action chunks with synchronized future views to model the visual consequences of the team's actions.

\paragraph{Predicting multi-agent worlds.}
Shared-world prediction provides a complementary perspective on collaboration. COMBO reconstructs shared state from egocentric observations for compositional planning, and SeqWM models multi-robot dynamics sequentially with intention sharing \citep{zhang2025combo,zhao2026empowering}. Solaris, ShareVerse, MultiWorld, Gamma-World, and population-scalable video models develop consistent multi-agent views and scalable video generation \citep{savva2026solaris,zhu2026shareverse,wu2026multiworld,liu2026gamma,zhao2026population}. Action-conditioned video models predict how supplied robot or player controls change the scene. For coordinated manipulation, the policy must also generate the controls that participate in that change. \method learns a joint distribution over the team's visual futures and low-level actions.

\paragraph{Learning actions with visual futures.}
Joint visual and action prediction connects generative world models to robot control. DreamZero learns aligned visual futures and robot actions; Motus unifies world modeling, policy learning, inverse dynamics, and joint video-action prediction, while Motubrain develops multiview and cross-embodiment modeling \citep{ye2026world,bi2026motus,team2026motubrain}. Recent work also clarifies how a policy can use future information. Fast-WAM and SelfWAM retain future supervision with action-only deployment paths, and RIFT constructs a future key/value interface in one backbone pass \citep{yuan2026fast,pan2026selfwam,zhang2026keep}. These designs connect a robot's actions to anticipated scene evolution. In a team, that evolution also depends on what the other robots do. \method extends joint world-action modeling to these cross-robot dependencies: all agents' action chunks and private-view futures interact within one generative process. Each robot's predicted behavior is thereby modeled alongside the future that the team jointly creates.

\section{\method}
\label{sec:method}

\begin{figure*}[ht!]
\centering
\includegraphics[width=\textwidth]{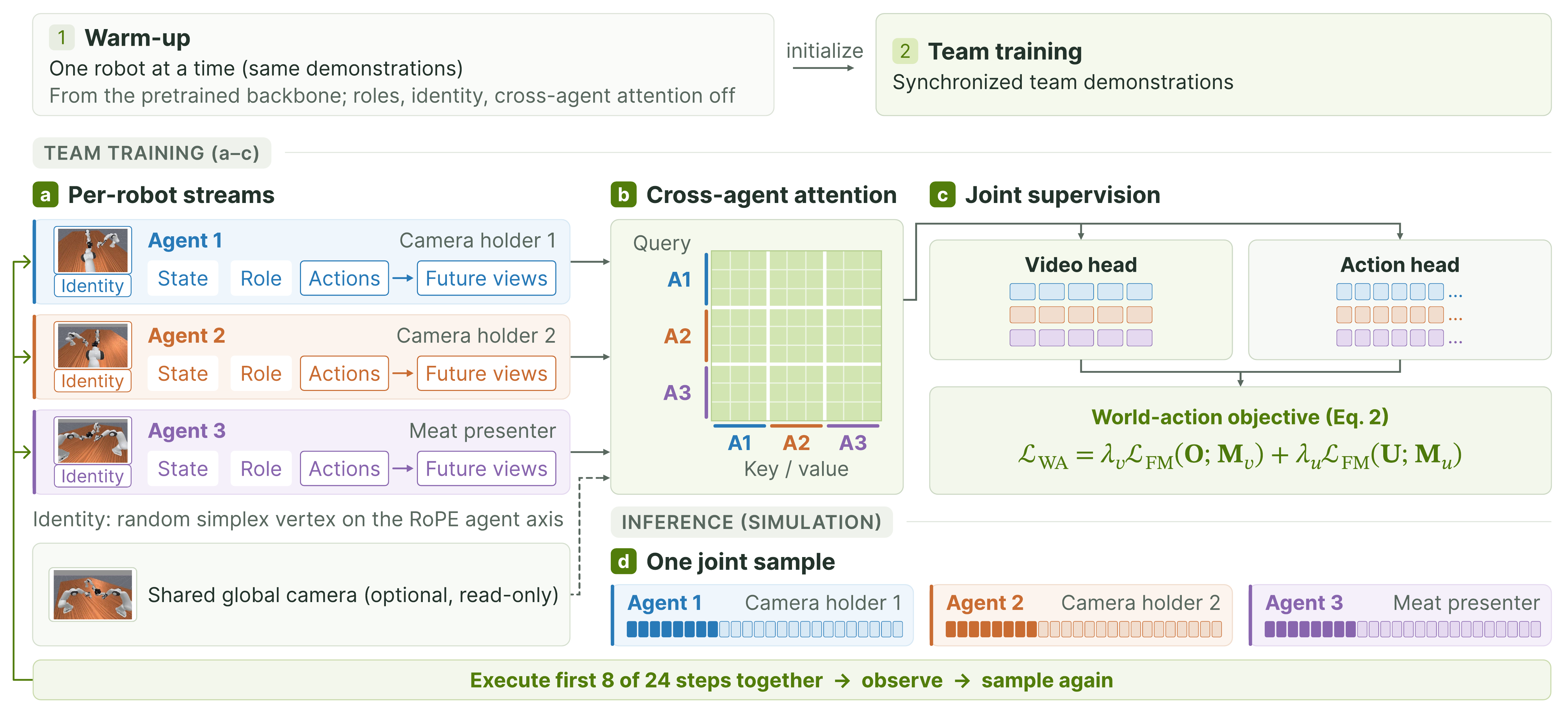}
\caption{\textbf{Joint actions, shared consequences.} Each robot's actions change the scene for its teammates. Warm-up adapts the pretrained single-robot backbone one robot at a time; team training (a--c) then learns from synchronized team demonstrations. (a)~Each robot's stream carries its views, state, role, and noisy action and future-view targets; a simplex identity on the RoPE agent axis marks its views, and an action-to-video bias (arrow) adds its action summary to its future views. (b)~One shared Transformer attends densely across all agents. (c)~Video and action heads share the world-action objective of Eq.~\eqref{eq:world-action-loss}. (d)~One joint sample gives every robot's action chunk and private-view future (Figure~\ref{fig:sim-qualitative}); the robots execute the first 8 of 24 steps together, observe, and replan. Shown: Camera Alignment in simulation, with one private view and a distinct role label per arm, plus a read-only shared global camera.}
\label{fig:method}
\end{figure*}

\method predicts each robot's actions together with the future the whole team will see. It extends a pretrained single-robot world-action backbone to generate the whole team's future views and action chunks in one sample (Figure~\ref{fig:method}). Cross-agent attention couples the robots' predictions; simplex identities attach every view to its robot, and an action-to-video bias and role embeddings attach its actions and task role.

\subsection{Predicting joint actions and their consequences}
\label{sec:joint-prediction}

Consider $N$ agents acting in one environment. An agent is one robot with its own cameras and controller: a dual-arm Galaxea R1 Pro in our real-robot experiments and a single Panda arm in simulation. Agent $i$ has state $\mathbf{s}^i_t$, action $\mathbf{u}^i_t$, task role $r_i$, and private camera streams $\mathcal{V}_i$. A view--agent map $q(v)=i$ assigns each stream to its owner; $c(v)$ specifies its camera type. The context $\mathcal{C}_t$ contains the instruction, embodiment metadata, observations, states, roles, and ownership labels, including optional shared global camera observations $\mathcal{G}$. We model
\begin{equation}
 p_\theta\!\left(
   \underbrace{
   \{\mathbf{o}^{v}_{t+3k}\}_{v\in\cup_i\mathcal{V}_i,\;k=1,\dots,8}
   }_{\mathbf{O}},
   \underbrace{
   \{\mathbf{u}^{i}_{t:t+H_u-1}\}_{i=1}^{N}
   }_{\mathbf{U}}
   \,\middle|\,\mathcal{C}_t
 \right).
 \label{eq:joint-distribution}
\end{equation}
Here, $H_u=24$ is the action horizon in control steps. Each view is observed at $t-12,t-9,\dots,t$ and predicted at $t+3k$ for $k=1,\dots,8$: eight future frames at a three-step interval that cover the same 24 steps (1.2\,s at 20\,Hz in simulation, 1.6\,s at 15\,Hz on the real robots). The future observations describe one unfolding interaction from the robots' different viewpoints; the action chunks specify how each robot participates in it. Supervising them jointly teaches the model the relationship between what the team does and what its cameras subsequently observe. Each action chunk is assigned to its controller. Private views provide the prediction targets; shared global cameras supply additional context.

\subsection{Coupling the team's predictions}
\label{sec:generator}

\paragraph{Cross-agent attention.}
One Transformer attends over all agents' view, state, and action tokens (\S~\ref{sec:ownership}) and jointly denoises their future views and actions. It uses dense bidirectional attention across agents and prediction time, so each agent's actions and future views are computed jointly with its teammates' tokens. In Long Pipeline Delivery, four arms relay a shoe down a line. Each arm can attend to its teammates' action and future-view tokens, allowing their handover motions to be generated jointly. The controls and visual consequences of all participants are generated in the same sample. Shared global camera tokens, when available, serve as read-only context.

\paragraph{Joint world-action learning.}
\label{sec:joint-objective}
Modality-specific heads predict private-view latents and action chunks using flow matching \citep{lipman2022flow}:
\begin{equation}
 \mathcal{L}_{\mathrm{WA}}
 =
 \lambda_v\mathcal{L}_{\mathrm{FM}}(\mathbf{O};\mathbf{M}_v)
 +
 \lambda_u\mathcal{L}_{\mathrm{FM}}(\mathbf{U};\mathbf{M}_u).
 \label{eq:world-action-loss}
\end{equation}
The visual mask $\mathbf{M}_v$ selects future targets from all valid private views, excluding observed, padded, and shared global camera tokens. The action mask $\mathbf{M}_u$ selects valid controller dimensions. Synchronized demonstrations supervise both parts of the joint sample. Appendix~\ref{app:flow-objective} specifies the forward process and normalization.

\subsection{Keeping each output with its robot}
\label{sec:ownership}

\paragraph{Per-robot streams.}
We group each robot's private views, state, and actions into one stream. The frozen video VAE encodes each private view, and its latent patches form spatiotemporal tokens $\mathcal{V}_{i,v}$; a learned view embedding, indexed by input position, distinguishes an agent's views. Encoders gather the valid state and action dimensions into fixed-width state and action tokens $\mathcal{S}_i$ and $\mathcal{U}_i$:
\begin{equation}
 \mathcal{X}_i =
 \left(\bigcup_{v\in\mathcal{V}_i}\mathcal{V}_{i,v}\right)
 \cup\mathcal{S}_i\cup\mathcal{U}_i .
 \label{eq:agent-stream}
\end{equation}
An action-to-video bias adds the mean of agent $i$'s encoded noisy action tokens to the future tokens of its private views, connecting control tokens to visual predictions. Each robot can have several private cameras but has a single state and action interface. A weight-shared decoder maps each agent's action tokens to its controller layout and returns the chunk to that controller. Appendix~\ref{app:ownership} defines the attribution operators; Appendices~\ref{app:simulation-layout} and~\ref{app:real-layout} specify controller layouts.

\paragraph{Identity and role.}
Joint prediction must distinguish the robots that act from the functions they perform. Identity groups one robot's views, and the action-to-video bias attaches its controls to them; role specifies its responsibility in the task. In Camera Alignment, each arm has its own role label: $[0,1,2]$ for the two camera holders and the meat presenter. If the two holders share one label ($[0,0,1]$), each keeps its own state and action tokens, and success is unchanged (Table~\ref{tab:ablations}). A role thus marks a common responsibility without merging the robots that fulfill it.
We add a learned role embedding $\rho(r_i)$ to agent $i$'s tokens and extend video RoPE on $(t,h,w)$ with an agent axis that takes the low-frequency half of the pretrained temporal band (Appendix~\ref{app:rotary}). Identity phases are drawn from $P$ vertices of a centered unit regular simplex:
\begin{equation}
 \mathbf{c}_j =
 \sqrt{\frac{P}{P-1}}
 \left(\mathbf{e}_j-\frac{1}{P}\mathbf{1}\right),
 \qquad
 \mathbf{c}_j^\top\mathbf{c}_k =
 \begin{cases}
 1, & j=k,\\[-2pt]
 -\frac{1}{P-1}, & j\ne k ,
 \end{cases}
 \label{eq:simplex}
\end{equation}
All views of an agent receive the same vertex; shared global camera tokens receive zero agent phase. The vertex enters RoPE as a constant rotation: the $P$ coordinates of $\mathbf{c}_j$ are the angles, in radians, of the first $P$ rotation pairs of the agent band, and the remaining agent pairs are not rotated. View tokens of agents $i$ and $j$ therefore differ by $\mathbf{c}_i-\mathbf{c}_j$ in the agent band at every time and pixel position; state and action tokens carry no agent phase (Appendix~\ref{app:rotary}). Equal pairwise distances give distinct identities a symmetric geometric code. During training, we randomly assign distinct vertices and shuffle complete agent groups, exposing each role to varied identity phases and sequence positions. The view embedding stays with the input position, so outputs are only approximately permutation-equivariant (Appendix~\ref{app:simplex-guarantees}).

\subsection{Training and closed-loop control}
\label{sec:training}

The model first learns how one robot acts through its own cameras and controller, then how the robots' actions interact. Both stages use Eq.~\eqref{eq:world-action-loss}.
\textbf{\textit{(1) \warmup.}}
Starting from a pretrained single-robot world-action backbone, the model learns from one robot's views, state, and actions at a time, with roles, identities, and cross-agent attention disabled. This adapts it to the cameras and controllers. In simulation, Warm-up also predicts the shared global camera like a private view.
\textbf{\textit{(2) \coordstage.}}
Initialized from the Warm-up weights, the model learns from synchronized team demonstrations with per-robot streams, roles, randomized identities, group shuffling, and cross-agent attention enabled.

\paragraph{Closed-loop joint control.}
At deployment, joint denoising produces all private-view futures and each robot's action chunk. Each controller executes its first $R$ actions in lockstep with the others. These actions change the scene for the whole team, and the updated observations and states condition the next joint sample.

\section{Experiments}
\label{sec:experiments}

We evaluate \method on two real robots and simulated teams with two to four arms, followed by component ablations.

\begin{table}[ht!]
\caption{\rev{\textbf{\method is best on every real-robot task.} Success rates (\%) with binomial standard error over 30 trials per task on two Galaxea R1 Pro robots. Laundry, Towel, and Umbrella denote laundry-bag collection, bath-towel folding, and umbrella handover; Avg.\ is the three-task mean. CLS-DP \citep{park2026distilling}, GauDP \citep{NEURIPS2025_0895de15}, and a $\pi_{0.5}$-based reimplementation of MA-VLA \citep{zhang2026ma} use the same demonstrations, robots, and three private views per robot, without a shared global camera. The last row is \method without cross-agent attention (independent agents).}}
\label{tab:real-results}
\centering
\scriptsize
\setlength{\tabcolsep}{5pt}
\resizebox{\columnwidth}{!}{%
\begin{tabular}{lcccc}
\toprule
Method & Laundry & Towel & Umbrella & Avg. \\
\midrule
CLS-DP
  & 23.3\sepm{7.7}
  & 33.3\sepm{8.6}
  & 56.7\sepm{9.0}
  & 37.8\sepm{4.9} \\
GauDP
  & \RealGauDPLaundry
  & \RealGauDPTowel
  & \RealGauDPUmbrella
  & \RealGauDPMacro \\
MA-VLA ($\pi_{0.5}$)
  & \second{\RealMAVLALaundry}
  & \second{\RealMAVLATowel}
  & \second{\RealMAVLAUmbrella}
  & \second{\RealMAVLAMacro} \\
\rowcolor{ourshade}\method (ours)
  & \best{\RealMawamLaundry}
  & \best{\RealMawamTowel}
  & \best{\RealMawamUmbrella}
  & \best{\RealMawamMacro} \\
\midrule
\method, independent agents
  & \RealIndepLaundry
  & \RealIndepTowel
  & \RealIndepUmbrella
  & \RealIndepMacro \\
\bottomrule
\end{tabular}}
\end{table}

\begin{table*}[ht!]
\caption{\rev{\textbf{Coordinating interdependent actions across two-, three-, and four-arm teams.} Success rates (\%) on eight \robofactory tasks; \method uses 100 evaluation episodes per task, with binomial standard errors. Team training uses 150 joint demonstrations per task. Bold and underline indicate the highest and second-highest reported rates.}}
\label{tab:sim-results}
\centering
\scriptsize
\setlength{\tabcolsep}{4pt}
\begin{tabularx}{\textwidth}{
  >{\raggedright\arraybackslash}p{3.3cm}
  c
  Y Y Y Y Y}
\toprule
& &
\shortstack{Decentralized policy\\coordination}
& \multicolumn{1}{c}{
    \shortstack{Joint trajectory\\and action generation}}
& \multicolumn{2}{c}{
    \shortstack{Language- and logic-guided\\task coordination}}
& \\
\cmidrule(lr){3-3}
\cmidrule(lr){4-4}
\cmidrule(lr){5-6}
Task & $N$
& CLS-DP
& GauDP
& LTLDiff
& MA-VLA
& \method (ours) \\
\midrule
Lift Barrier & 2
& \CLSDPLift
& \GauDPLift
& \LTLDiffLift
& \second{\MAVLALift}
& \cellcolor{ourshade}\best{\SimMawamLiftPct} \\

Pass Shoe & 2
& \CLSDPPassShoe
& \GauDPPassShoe
& \LTLDiffPassShoe
& \second{\MAVLAPassShoe}
& \cellcolor{ourshade}\best{\SimMawamPassShoePct} \\

Place Food & 2
& \CLSDPPlaceFood
& \GauDPPlaceFood
& \LTLDiffPlaceFood
& \second{\MAVLAPlaceFood}
& \cellcolor{ourshade}\best{\SimMawamPlaceFoodPct} \\

Two Robots Stack Cube & 2
& \CLSDPTwoStack
& \GauDPTwoStack
& \LTLDiffTwoStack
& \second{\MAVLATwoStack}
& \cellcolor{ourshade}\best{\SimMawamTwoStackPct} \\

Camera Alignment & 3
& \CLSDPCameraAlign
& \GauDPCameraAlign
& \LTLDiffCameraAlign
& \second{\MAVLACameraAlign}
& \cellcolor{ourshade}\best{\SimMawamCameraAlignPct} \\

Three Robots Stack Cube & 3
& \CLSDPThreeStack
& \GauDPThreeStack
& \LTLDiffThreeStack
& \second{\MAVLAThreeStack}
& \cellcolor{ourshade}\best{\SimMawamThreeStackPct} \\

Take Photo & 4
& \CLSDPTakePhoto
& \GauDPTakePhoto
& \LTLDiffTakePhoto
& \second{\MAVLATakePhoto}
& \cellcolor{ourshade}\best{\SimMawamTakePhotoPct} \\

Long Pipeline Delivery & 4
& \CLSDPPipeline
& \GauDPPipeline
& \LTLDiffPipeline
& \best{\MAVLAPipeline}
& \cellcolor{ourshade}\second{\SimMawamPipelinePct} \\

\midrule
\textbf{Average over the eight tasks} &
& \CLSDPMacro
& \GauDPMacro
& \LTLDiffMacro
& \second{\MAVLAMacro}
& \cellcolor{ourshade}\best{\SimMawamMacro} \\
\bottomrule
\end{tabularx}
\end{table*}
\begin{figure*}[ht!]
\centering
\includegraphics[width=\textwidth]{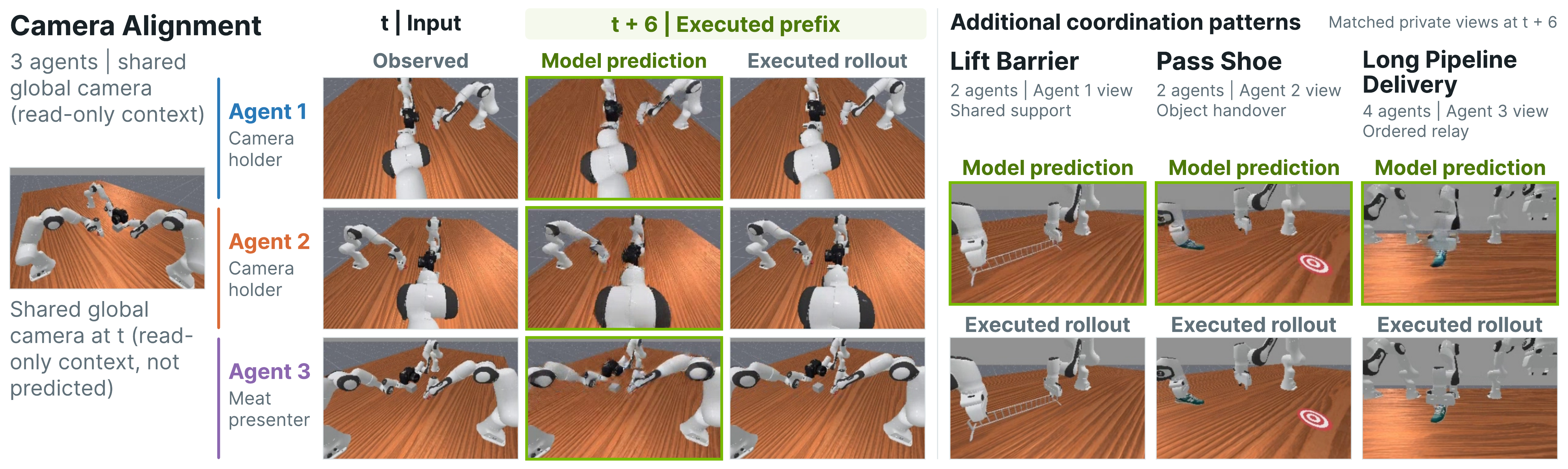}
\caption{\textbf{Predicted and executed observations.} Camera Alignment in simulation: for each agent (two camera holders and a meat presenter), the observed private view at $t$, \method's prediction six control steps ahead from one joint sample, and the executed view at the same step. The team executes the first eight steps of its 24-step action chunks before replanning; the shared global camera is read-only context and is not predicted. Right: matched predicted and executed private views for Lift Barrier, Pass Shoe, and Long Pipeline Delivery. Quantitative evaluation in Appendix~\ref{app:video-accuracy}.}
\label{fig:sim-qualitative}
\end{figure*}

\subsection{Real-world collaboration from private views}
\label{sec:real}

\paragraph{Tasks and platform.}
We evaluate three collaborative tasks using two Galaxea R1 Pro robots: laundry-bag collection, bath-towel folding, and umbrella handover. In laundry-bag collection, both robots must keep the bag lifted and open while a garment is placed inside; success requires the garment to be inside the bag at termination, with neither robot grasping it. Bath-towel folding requires coordinated manipulation of a deformable object, whereas umbrella handover requires transferring support from one robot to the other. Each robot has one ego/head-left camera and two wrist cameras. The model therefore receives six private camera views, with no shared global camera, and must integrate their complementary observations to predict the team's future observations and actions.

\paragraph{Training and evaluation.}
We initialize \method from GR00T~2 and use \warmup to adapt it to each robot's cameras and controller. We then train on synchronized demonstrations of the robot pair, jointly supervising actions and future observations in all six camera views. Umbrella handover uses \textsc{giver}/\textsc{receiver} roles; laundry and towel roles follow the collection setup. Demonstrations are split at the synchronized-pair level after camera, timing, and action checks. At deployment, the robots execute the first 8 of the 24 predicted steps together, observe the resulting scene, and replan.
We compare \method with CLS-DP \citep{park2026distilling}, GauDP \citep{NEURIPS2025_0895de15}, and a $\pi_{0.5}$-based reimplementation of MA-VLA \citep{zhang2026ma}. All methods use the same demonstrations, robots, and private-camera setup. Each method is evaluated over 30 trials per task. The robots start from a common fixed initial pose (Appendix Table~\ref{tab:real-home}), while object placement and state are randomized within bounded ranges. Success is assessed using task-specific terminal criteria; interrupted episodes and episodes with human intervention count as failures. For each task, we report the success rate and its binomial standard error $\sqrt{p(1-p)/n}$. Deployment and evaluation details are provided in Appendix~\ref{app:real}.

\paragraph{Results.}
\method achieves the highest success rate on all three tasks (Table~\ref{tab:real-results}), averaging 67.8\%, compared with 48.9\% for MA-VLA, 37.8\% for CLS-DP, and 28.9\% for GauDP. The largest improvement over MA-VLA occurs on laundry-bag collection, where success increases from 30.0\% to 53.3\%. Removing cross-agent attention (independent agents) lowers \method's mean success to 52.2\%, with the largest drop on laundry-bag collection (53.3\% to 36.7\%), where loading the garment depends on both robots holding the bag open.

\newsavebox{\abPanelB}
\begin{table*}[ht!]
\caption{\textbf{Ablation study of joint prediction and coordinated execution.} Success rates (\%) with binomial standard error over 100 episodes per task. Each panel shows the tasks its variants were evaluated on (--: not evaluated); the shaded row is the full model. (a)~\emph{Independent agents} removes cross-agent attention; \emph{No team-stage future-video supervision} keeps the future-view tokens and inference; \emph{No shared global camera} drops the shared global camera in team training and evaluation; \emph{No Warm-up} starts team training from the pretrained backbone. (b)~One role for all agents (\emph{No roles}) or for the two camera holders only (\emph{Shared-holder roles}). (c)~\emph{Absolute slots} (learned slot IDs) or \emph{Fixed simplex} vertices, both with a fixed assignment, replace the randomized simplex identities. (d)~\emph{Centralized team} keeps the future-video loss and the shared global camera but uses one visual context and one action vector for all arms, with an interface warm-up in place of Warm-up; each arm's fixed position sets its identity and role (Appendix~\ref{app:ablations}).}
\label{tab:ablations}
\label{tab:role-diagnostics}
\centering
\scriptsize
\newcommand{\abhead}[2]{\begin{tabular}[b]{@{}c@{}}#1\\#2\end{tabular}}
\setlength{\tabcolsep}{3pt}
\sbox{\abPanelB}{\begin{tabular}[t]{lcc}
\multicolumn{3}{l}{\small (b) Task roles}\\[3pt]
\toprule
Variant & \abhead{Place}{Food} & \abhead{Camera}{Alignment} \\
\midrule
\rowcolor{ourshade}\method{} (full) & \SimMawamPlaceFoodPct & \SimMawamCameraAlignPct \\
No roles & \SimNoRolesPlaceFood & \SimNoRolesCameraAlign \\
Shared-holder roles & n/a & \SimSharedHolderCameraAlign \\
\bottomrule
\end{tabular}}%
{\renewcommand{\tabularxcolumn}[1]{m{#1}}%
\begin{tabularx}{\dimexpr\textwidth-\wd\abPanelB-1em\relax}[t]{>{\raggedright\arraybackslash}Xcccccc}
\multicolumn{7}{l}{\small (a) Cross-agent coupling, future prediction, information access, and training}\\[3pt]
\toprule
Variant & \abhead{Lift}{Barrier} & \abhead{Place}{Food} & \abhead{Camera}{Alignment} & \abhead{Two Robots}{Stack Cube} & \abhead{Pass}{Shoe} & \abhead{Long Pipeline}{Delivery} \\
\midrule
\rowcolor{ourshade}\method{} (full)
  & \SimMawamLiftPct & \SimMawamPlaceFoodPct & \SimMawamCameraAlignPct & \SimMawamTwoStackPct & \SimMawamPassShoePct & \SimMawamPipelinePct \\
Independent agents
  & \SimIndependentLift & \SimIndependentPlaceFood & \SimIndependentCameraAlign & -- & \SimIndependentPassShoe & \SimIndependentPipeline \\
No team-stage future-video supervision
  & \SimPerAgentActionLift & \SimPerAgentActionPlaceFood & \SimPerAgentActionCameraAlign & \SimPerAgentActionTwoStack & -- & \SimPerAgentActionPipeline \\
No shared global camera
  & \SimNoGlobalLift & \SimNoGlobalPlaceFood & \SimNoGlobalCameraAlign & \SimNoGlobalTwoStack & -- & -- \\
No Warm-up
  & \SimNoWarmupLift & \SimNoWarmupPlaceFood & \SimNoWarmupCameraAlign & -- & \SimNoWarmupPassShoe & \SimNoWarmupPipeline \\
\bottomrule
\end{tabularx}}\hfill
\usebox{\abPanelB}

\vspace{8pt}
\begin{tabularx}{0.485\textwidth}[t]{lYYYY}
\multicolumn{5}{l}{\small (c) Agent identity}\\[3pt]
\toprule
Variant & \abhead{Lift}{Barrier} & \abhead{Place}{Food} & \abhead{Camera}{Alignment} & \abhead{Take}{Photo} \\
\midrule
\rowcolor{ourshade}\method{} (full) & \SimMawamLiftPct & \SimMawamPlaceFoodPct & \SimMawamCameraAlignPct & \SimMawamTakePhotoPct \\
Absolute slots & \SimAbsSlotsLift & \SimAbsSlotsPlaceFood & \SimAbsSlotsCameraAlign & \SimAbsSlotsTakePhoto \\
Fixed simplex & \SimFixedSimplexLift & \SimFixedSimplexPlaceFood & \SimFixedSimplexCameraAlign & \SimFixedSimplexTakePhoto \\
\bottomrule
\end{tabularx}\hfill
\begin{tabularx}{0.485\textwidth}[t]{lYYY}
\multicolumn{4}{l}{\small (d) Baseline: Centralized team}\\[3pt]
\toprule
Variant & \abhead{Place}{Food} & \abhead{Pass}{Shoe} & \abhead{Long Pipeline}{Delivery} \\
\midrule
\rowcolor{ourshade}\method{} (full) & \SimMawamPlaceFoodPct & \SimMawamPassShoePct & \SimMawamPipelinePct \\
Centralized team & \SimCentralPlaceFoodPct & \SimCentralPassShoePct & \SimCentralPipelinePct \\
\bottomrule
\end{tabularx}
\end{table*}

\subsection{Simulation with two- to four-arm teams}
\label{sec:sim}
\begingroup\sepmtablesonly  

We evaluate all eight multi-robot \robofactory tasks \citep{qin2025robofactory}, spanning two, three, and four arms. Camera Alignment couples a camera-holder pair with a meat presenter that holds the target up to the camera; Take Photo adds a fourth arm responsible for triggering the camera. Long Pipeline Delivery connects four arms through successive handovers. \method is initialized from GR00T~2, with 150 synchronized demonstrations per task for team training. Each replan predicts a 24-step action chunk for every agent together with its future views; the agents execute the first eight steps in lockstep and then replan from live observations (Figure~\ref{fig:sim-qualitative}).
Each task is evaluated on 100 episodes. Appendix Table~\ref{tab:task-taxonomy} describes each task's roles. Table~\ref{tab:sim-results} includes results for GauDP, CLS-DP, LTLDiff, and MA-VLA.
\method achieves an average success rate of \SimMawamMacro\% across the eight tasks (Table~\ref{tab:sim-results}). On the two four-arm tasks, Take Photo and Long Pipeline Delivery, it reaches \SimMawamTakePhotoPct\% and \SimMawamPipelinePct\%, respectively, covering both joint manipulation and sequential handovers. The largest gains over the strongest published baseline occur on Place Food (\SimMawamPlaceFoodPct\% vs.\ \MAVLAPlaceFood\%) and Take Photo (\SimMawamTakePhotoPct\% vs.\ \MAVLATakePhoto\%).

\subsection{Ablation study}
\label{sec:ablations}

We examine, in order, coupling between agents, future prediction, information access and training, and agent representation. Table~\ref{tab:ablations} follows the same order, and Appendix~\ref{app:ablations} specifies each configuration.

\paragraph{Cross-agent attention.}
Disabling cross-agent attention reduces success on Long Pipeline Delivery from \SimMawamPipelinePct\% to \SimIndependentPipeline\% and on Pass Shoe from \SimMawamPassShoePct\% to \SimIndependentPassShoe\%. Each agent retains access to its private views and the read-only shared global camera but cannot attend to other agents' tokens. With the backbone, parameter count, demonstrations, and training budget fixed, these results support the value of cross-agent attention for coordinated handovers; on the real robots the same ablation lowers mean success from 67.8\% to 52.2\% (Table~\ref{tab:real-results}).

\begin{figure}[ht!]
\centering
\includegraphics[width=\columnwidth]{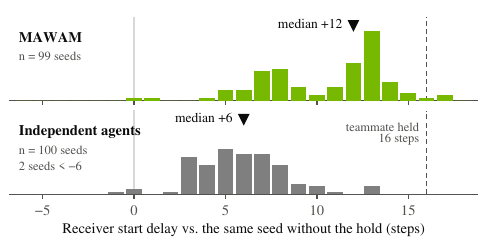}
\caption{\textbf{Response to a delayed teammate.} In Long Pipeline Delivery, we hold the giver in the second handover stationary for 16 control steps after it releases the shoe. For each seed, we measure the receiver's start delay relative to the same seed without the hold.}
\label{fig:teammate-delay}
\end{figure}

\paragraph{Response to a delayed teammate.}
To test whether the receiving agent adjusts its timing to a teammate, we introduce a 16-step hold in Long Pipeline Delivery: the giver in the second handover is held stationary after releasing the shoe (100 paired seeds). \method delays the receiver's start by a median of 12 steps; independent agents show a median delay of 6 steps (Figure~\ref{fig:teammate-delay}). \method's delay carries through to the grasp (median $+9$ vs.\ $+1.5$ steps) and is still 10 steps without the 18 episodes in which its giver re-grasps; under the hold, \method succeeds in 99\% of episodes vs.\ 44\% for independent agents (Appendix~\ref{app:ablations}).

\paragraph{Future-video supervision and tokens.}
Removing the future-video loss in team training, with the same future-view tokens, attention, and inference computation, reduces success on Place Food from \SimMawamPlaceFoodPct\% to \SimPerAgentActionPlaceFood\% and on Camera Alignment from \SimMawamCameraAlignPct\% to \SimPerAgentActionCameraAlign\%. Team-stage future-video supervision thus helps most where arms must hold a shared configuration, such as keeping a lid open while another arm places the meat, or keeping the camera aligned with the meat the presenter raises. Where the scene changes, the predicted views are 4--7\,dB closer (PSNR) to the executed frames than a copy of the last frame (Appendix Table~\ref{tab:video-accuracy}).
At test time, retaining the future-view tokens without denoising them yields similar success rates across all eight tasks (Appendix Table~\ref{tab:test-time-video}). The model is trained with these tokens in every sample, and removing them at test time lowers success on Long Pipeline Delivery from \SimMawamPipelinePct\% to \SimTTNoFutPipeline\%, on Pass Shoe from \SimMawamPassShoePct\% to \SimTTNoFutPassShoe\%, and on Two Robots Stack Cube from \SimMawamTwoStackPct\% to \SimTTNoFutTwoStack\%. This removal reduces model inference time on the real-robot setup from 0.32\,s to 0.22\,s. The future-view tokens are thus the shared space where the team's actions are computed, and control does not need a finished video, consistent with recent single-robot findings that world-action models use their future tokens without test-time imagination~\citep{yuan2026fast,pan2026selfwam}.

\paragraph{Private and shared observations.}
Removing the shared global camera during team training and evaluation leaves each agent with only its private view. Success rates remain within 4 percentage points of the full model on Lift Barrier, Camera Alignment, and Two Robots Stack Cube. Place Food is more sensitive: success falls from \SimMawamPlaceFoodPct\% to \SimNoGlobalPlaceFood\% without the shared global camera, which is the only view that captures both objects held by the arms. Together with the real-world experiments, these results show that private observations can support joint control, while a shared view can help when it captures task-relevant relationships missing from individual views.

\paragraph{Warm-up.}
Warm-up improves performance compared with starting team training directly from the pretrained backbone. Its largest gain occurs on Long Pipeline Delivery, where success increases from \SimNoWarmupPipeline\% to \SimMawamPipelinePct\%. It also improves Place Food from \SimNoWarmupPlaceFood\% to \SimMawamPlaceFoodPct\%, with smaller gains on Camera Alignment and Pass Shoe.

\paragraph{Task roles and agent identity.}
Assigning the same role to every agent reduces success on Camera Alignment from \SimMawamCameraAlignPct\% to \SimNoRolesCameraAlign\% and on Place Food from \SimMawamPlaceFoodPct\% to \SimNoRolesPlaceFood\%. In Camera Alignment, this variant lifts the camera in 74 of 100 episodes but lifts the meat in only 8. Giving the two camera holders a shared role while retaining a separate role for the meat presenter achieves \SimSharedHolderCameraAlign\%. Thus, agents can share a task role while retaining distinct observations and controls.
We also compare randomized simplex identities with learned absolute slot IDs and fixed simplex vertex assignments. The largest benefit appears on Place Food, where the three variants achieve \SimMawamPlaceFoodPct\%, \SimAbsSlotsPlaceFood\%, and \SimFixedSimplexPlaceFood\%, respectively. The improvement is smaller on Lift Barrier. Take Photo success rates are \SimMawamTakePhotoPct\%, \SimAbsSlotsTakePhoto\%, and \SimFixedSimplexTakePhoto\%.

\paragraph{Centralized team.}
The \emph{Centralized team} baseline treats the team as one robot, with one visual context, one action vector, and no per-robot tokens, identity codes, or role embeddings. Each arm's fixed position still encodes its role, as no arm switches sub-tasks across demonstrations. With the same backbone, demonstrations, and team-stage training budget, it trails \method on Pass Shoe and Place Food and fails on four-arm Long Pipeline Delivery (\SimCentralPipelinePct\% vs.\ \SimMawamPipelinePct\%), where no arm grasps the shoe even after extended interface warm-up (Appendix~\ref{app:ablations}). GR00T~2 was pretrained on single-agent inputs and outputs, which the Centralized team must relearn in a merged four-arm format, whereas \method keeps every robot in the pretrained format and so benefits directly from that pretraining.

\endgroup

\section{Conclusion}
\label{sec:conclusion}

A robot team's actions jointly create the world in which each member must act next. \method generates every robot's next action chunk and future camera views in one sample. On two real robots that see the scene only through their own cameras, it is best on all three tasks with 67.8\% mean success, 18.9 percentage points above MA-VLA; in simulation, it averages 90.3\% on eight tasks with two to four arms. Cross-agent attention and team-stage future-video supervision both improve coordinated control: without attention, a four-arm relay falls from 95.0\% to 39.0\% and real-robot mean success from 67.8\% to 52.2\%; without this supervision, Place Food falls from 78.0\% to 47.0\%. Training one model to predict each robot's next moves together with the future those moves create is a practical basis for coordinated manipulation.

\section{Limitations and Future Work}
Our experiments are limited to fixed robot teams of up to four agents. One joint sample serves the whole team, and going from two to four arms raises its time only from 1.40\,s to 1.85\,s (Appendix Table~\ref{tab:sim-latency}). Future work could extend \method to larger teams, allow team membership to change during execution, and support collaboration among robots with different embodiments and sensors. The predicted future observations and actions could help people understand a team's intended behavior before it acts. Extending joint prediction to include human partners could also help robots anticipate human actions and coordinate with them.

\appendix
\newpage
\section*{Appendix}
The appendices give the details behind the main paper. Figure, table, and equation numbers continue from the main paper, and references to sections, tables, and equations without the word ``Appendix'' point to the main paper. Appendix~\ref{app:symmetry} defines the functional roles and task symmetries of the eight \robofactory tasks (Table~\ref{tab:task-taxonomy}), shows how each private view and action coordinate is routed to its robot, and states the permutation properties of the simplex identity codes. Appendix~\ref{app:implementation} gives the implementation: the action and camera layouts in simulation and on the real robots, the rotary-band allocation, the Warm-up and team-training procedure and configurations (Tables~\ref{alg:training} and~\ref{tab:stage-config}), and the flow schedules and masked training objectives. Appendix~\ref{app:evaluation} defines simulation success and its standard error, and measures how closely the predicted views match the executed frames (Table~\ref{tab:video-accuracy}). Appendix~\ref{app:ablations} specifies every ablation in Section~\ref{sec:ablations}, including the test-time video controls (Table~\ref{tab:test-time-video}), the centralized-team baseline, and the teammate hold. Appendix~\ref{app:real} covers the real robots: hardware, initial poses, cameras, demonstration collection, post-training and deployment, success and safety accounting, and model size and inference time (Tables~\ref{tab:real-efficiency} and~\ref{tab:sim-latency}).

\section{Task symmetry and agent equivariance}
\label{app:symmetry}

\paragraph{Role graphs.}
\label{app:role-graphs}

A task's role graph records how agents depend on one another through precedence, object transfer, and shared-object constraints. We use a directed labeled graph with one node per agent slot, task roles as node labels, and these dependencies as labeled edges. An agent permutation is a task symmetry if it preserves node labels, edge labels, and embodiment metadata. This gives three team structures:

\begin{enumerate}
 \item \textbf{fully symmetric teams}, with one role class and admissible group $S_N$;
 \item \textbf{asymmetric teams}, whose role-preserving group may contain only the identity; and
 \item \textbf{mixed teams}, with role classes of sizes $n_1,\ldots,n_J$.
\end{enumerate}
The role-preserving task-symmetry group is a subgroup of $\prod_j S_{n_j}$; equality holds when all within-role permutations preserve the dependency graph and embodiment metadata.
Table~\ref{tab:task-taxonomy} summarizes the categorization of the \robofactory tasks used in our experiments.

\begin{table}[ht!]
\caption{\rev{\textbf{Task roles in the eight \robofactory tasks.} The tasks span shared-object support, asymmetric transfers, and ordered relays. Symmetry is defined over the whole-episode role graph and is a property of the task, not of the role labels given to the model: the main configurations of Camera Alignment and Take Photo give each camera holder its own label, and grouping the two Camera Alignment holders under one label is evaluated in Table~\ref{tab:role-diagnostics}.}}
\label{tab:task-taxonomy}
\centering
\scriptsize
\setlength{\tabcolsep}{3pt}
\begin{tabularx}{\columnwidth}{>{\raggedright\arraybackslash}p{2.35cm}ccX}
\toprule
Task & $N$ & Symmetry & Role structure \\
\midrule
Pass Shoe                  & 2 & \asym  & giver; receiver \\
Place Food                 & 2 & \asym  & meat handler; lid holder \\
Lift Barrier               & 2 & \sym   & exchangeable barrier holders, $S_2$ \\
Two Robots Stack Cube      & 2 & \asym  & shared pickup motif; then base/top roles \\
\midrule
Camera Alignment           & 3 & \mixed & camera-holder pair, $S_2$; meat presenter \\
Three Robots Stack Cube    & 3 & \asym  & shared pickup motif; then base/middle/top roles \\
\midrule
Take Photo                 & 4 & \mixed & camera-holder pair, $S_2$; object holder; trigger \\
Long Pipeline Delivery     & 4 & \asym  & repeated handover motif; ordered relay stations \\
\bottomrule
\end{tabularx}
\end{table}

\subsection{\rev{View ownership and action routing}}
\label{app:ownership}

Joint prediction retains an explicit correspondence between each robot's views and its action tokens. Let $V$ be the number of private views in one batch item. In the notation of Section~\ref{sec:method}, the view--agent map is $\mathbf q\in\{1,\ldots,N\}^{V}$; the converter serializes its zero-based equivalent
\begin{equation}
 \bar{\mathbf{q}}=\mathbf q-\mathbf 1\in\{0,\ldots,N-1\}^{V},
 \qquad
 \mathbf{c}\in\{1,\ldots,C_{\mathrm{type}}\}^{V},
\end{equation}
where $\bar q_v$ is the serialized owning-agent ID and $c_v$ is the camera type.  These arrays are batch-uniform after padding, and padded views carry a false validity mask.  The representation requires:
\begin{enumerate}
 \item every valid private view has exactly one valid agent owner;
 \item every agent with action tokens owns at least one valid view;
 \item the number of action and state agents is identical;
 \item roles are indexed by semantic agent, not by camera;
 \item camera-type labels are preserved under agent permutation; and
 \item shared global cameras, if present, are represented separately and never assigned a private agent ID.
\end{enumerate}
Samples with a missing view--agent map, inconsistent role count, unequal token width, or unknown camera type are excluded during preprocessing. Agent identity is read from explicit metadata rather than inferred from camera names or ordering.

\paragraph{Controller attribution.}
\label{app:action-scatter}
Let $A_i\in\{0,1\}^{d_i\times D}$ select agent $i$'s valid coordinates from a physical team-action vector $\mathbf{u}\in\mathbb{R}^{D}$.  The compact target is $\mathbf{u}^i=A_i\mathbf{u}$, and the decoded joint action is
\begin{equation}
 \widehat{\mathbf{u}}
 =\sum_{i=1}^{N}A_i^\top D_\theta(\mathcal{U}_i).
 \label{eq:action-scatter}
\end{equation}
The selection matrices have disjoint destinations.  Compact slices are padded to the common token width when needed, and validity masks exclude padded or unavailable coordinates from both decoding and loss computation.

For \robofactory, the map is one private view per Panda plus one optional shared global camera.  For the real setup,
\begin{gather*}
 \bar{\mathbf{q}}=[0,0,0,1,1,1],\\
 \mathbf{c}=[c_{\mathrm{ego}},c_{\mathrm{left}},c_{\mathrm{right}},
             c_{\mathrm{ego}},c_{\mathrm{left}},c_{\mathrm{right}}],
\end{gather*}
with no shared view.  Here $0$ and $1$ are semantic agent indices, not global camera IDs.

Camera and base calibration are used by the external data and control interfaces for stream association and coordinate conversion; they are not provided as Transformer inputs.

\subsection{\rev{Agent identity and permutation properties}}
\label{app:simplex-guarantees}

Let $\mathbf{B}=[\mathbf{c}_1,\ldots,\mathbf{c}_P]^\top$ contain the regular-simplex vertices from Equation~\ref{eq:simplex}.  For any vertex-permutation matrix $P_\pi$,
\begin{equation}
 (P_\pi \mathbf{B})(P_\pi \mathbf{B})^\top
 =P_\pi \mathbf{B}\mathbf{B}^\top P_\pi^\top
 =\mathbf{B}\mathbf{B}^\top .
\end{equation}
Thus simplex coordinates impose no distance-based order among agents.

We distinguish two tests.  \emph{Stream reordering} moves an agent's assigned phase, all private views, camera types, role, state and action tokens, diffusion noise, and scatter map together.  Write the model output as $f_\theta(\mathcal{C},\mathbf{z},\phi)$, where $\mathcal{C}$ is the context $\mathcal{C}_t$ of Eq.~\eqref{eq:joint-distribution}, $\mathbf{z}$ is the diffusion noise on all future-view and action tokens, and $\phi$ assigns a simplex vertex to each agent; $f_u$ and $f_v$ denote its action and future-video parts.  For any legal complete-group permutation $\pi$, shared weights and an index-equivariant mask \rev{make every operation commute with $\pi$ except the view embedding, which stays with the input position. The model is therefore only approximately equivariant:}
\begin{equation}
 f_\theta(\pi\!\cdot\!\mathcal{C},
          \pi\!\cdot\!\mathbf{z},
          \pi\!\cdot\!\phi)
 \approx
 \pi\!\cdot f_\theta(\mathcal{C},\mathbf{z},\phi).
 \label{eq:equivariance}
\end{equation}
\rev{Training with shuffled complete groups encourages this property.}
\emph{Vertex reassignment} is a different operation: it keeps each physical agent's observations, noise, and output order fixed and changes only its identity code $\phi$, the simplex vertex assigned to its views. A role-only swap changes the task metadata while leaving observation and controller ownership fixed; Appendix~\ref{app:ablations} reports its results.

\section{Implementation details}
\label{app:implementation}

\subsection{Simulation action and camera layout}
\label{app:simulation-layout}

Each Panda is represented by an equal-width 17D compact slice
\begin{equation}
 [\,\mathrm{EEF}_{9\mathrm{D}},
    \mathrm{gripper}_{1\mathrm{D}},
    \mathrm{joint}_{7\mathrm{D}}\,].
\end{equation}
State and action targets use the same pose convention.  Place Food gathers two slices into a $34$D joint tensor; Camera Alignment gathers three into $51$D. The model converts these source ranges to identically shaped per-agent tokens and inverse-scatters decoded tokens to the original controller layout.  The simulator executes the joint and gripper coordinates under position control; EEF coordinates remain supervised auxiliary components.

RoboFactory inputs are ordered as the optional shared global camera followed by one private view per Panda.  \rev{In team training,} the shared global camera uses current-frame repetition, read-only attention, zero future-video loss, and whole-stream conditioning dropout.  Private views carry future targets and nonzero video loss.  \rev{In Warm-up, each sample holds the shared global camera and one Panda's private view, and both are predicted and supervised.}  Camera order, source-arm order, and action destination are validated against episode metadata before training and evaluation.

Every view is sampled at offsets $-12,-9,\dots,24$ control steps from the current step $t$: five observed frames spanning 0.6\,s and eight future frames at a 0.15\,s interval up to $t+24$ (1.2\,s) at the 20\,Hz recording rate.  The action chunk covers steps $t$ to $t+23$, and the state input is step $t$.  The causal Wan~2.1 video VAE compresses time by $4\times$ and gives four latent frames per view, covering $\{t-12\}$, $\{t-9,\dots,t\}$, $\{t+3,\dots,t+12\}$, and $\{t+15,\dots,t+24\}$.  The first two are clean conditioning; the last two are generated for each private view.  \rev{In team training,} the shared global camera repeats its current latent in all four slots and is not predicted.  At test time, the history buffer is initialized by repeating the first observation, and the team executes the first 8 steps (0.4\,s) of each chunk before replanning.

\subsection{Real-robot observation and control layout}
\label{app:real-layout}

Each R1 Pro contributes ego/head-left, left-wrist, and right-wrist video. Mounted cameras are agent-local even when they show both robots.  There are six private views, zero shared views, two semantic agents, and two assigned simplex vertices sampled from the four-vertex pool described below.  All three views of one robot use the same agent phase and role while retaining different \rev{view embeddings, one per input position, initialized from the Warm-up embeddings of the ego, left-wrist, and right-wrist views.}  The action-to-video bias and the role embedding are broadcast to all three owned views.

The physical interface exposes 44 state and 37 command coordinates per robot. The learned action tokens activate 16 logical coordinates: left/right 7D relative end-effector commands and two absolute grippers.  The two-agent physical packet is therefore 88D state / 74D command, while the learned active space is $2\times16$D.  The action adapter writes inactive physical fields from the live controller state. The reported real-robot policy predicts 24-step action chunks, executes the first 8 steps, and replans from live observations. Frames use the same offsets as in simulation: five observed frames spanning the last 0.8\,s and eight future frames at $t+3,\dots,t+24$, a 0.2\,s interval and a 1.6\,s horizon at 15\,Hz.  This gives two generated latent frames per view and twelve across the six views.  The last future frame is the observation after the final step of the predicted chunk, so the visual and action horizons cover the same 1.6\,s.

\subsection{Rotary bands}
\label{app:rotary}

For a 128D attention head, the implementation allocates
\begin{equation}
 (d_t,d_a,d_h,d_w)=(22,22,42,42)
\end{equation}
rotary dimensions to time, agent, height, and width.  The 64 rotation angles of a head are ordered as 11 time, 11 agent, 21 height, and 21 width pairs, and, following the Cosmos-Predict2.5 convention, dimension $k$ is rotated together with dimension $k+64$; the agent band thus occupies dimensions 11--21 and 75--85.  Time, height, and width use the frequencies $\omega_k=10000^{-2k/d}$ with $d=d_t$, $d_h$, $d_w$ respectively.  Time positions are the latent-frame indices $0,\dots,3$ of each view, and height and width positions are latent patch indices.  The agent band takes the 11 lowest-frequency temporal pairs of the GR00T~2 attention head ($10^{-2}$ to $1.5\times10^{-4}$\,rad per latent frame).  Over the four latent frames of a view these pairs rotate by at most 0.03\,rad, so they carry almost no temporal information; the seven agent pairs without simplex coordinates stay unrotated.  The remaining 11 temporal pairs use every second frequency of the pretrained ladder, and the height and width bands drop the pretrained NTK rescaling (scale 3).  Warm-up keeps the pretrained rotary embedding; team training introduces this layout.

The agent band has no frequency ladder.  A pool of $P=4$ vertices $\mathbf{c}_j\in\mathbb{R}^4$ (Eq.~\eqref{eq:simplex}) supports the two-, three-, and four-agent simulation tasks and the two-agent real setup.  If agent $i$ is assigned vertex $\mathbf{c}_{\pi(i)}$, agent pair $k$ of every token in its views is rotated by $\alpha[\mathbf{c}_{\pi(i)}]_k$ for $k<4$ and by $0$ for $4\le k<11$, with $\alpha=1$.  Each vertex has $\sqrt{3}/2$ on its own coordinate and $-1/(2\sqrt{3})$ on the other three, so in $\mathbf{q}^\top\mathbf{k}$ the relative agent rotation is zero within an agent and $\pm 2/\sqrt{3}\approx 1.15$\,rad on pairs $\pi(i)$ and $\pi(j)$ between agents $i$ and $j$.  During training, $\pi$ maps the $N$ agents to $N$ distinct vertices drawn uniformly from the pool once per training step; at inference, agent $i$ uses vertex $i$.  Multiple views owned by one agent reuse its phase rather than consuming additional simplex vertices.  Shared global camera and padding tokens have zero agent phase.  State and action tokens use a one-dimensional rotary embedding over the chunk step, identical for all agents, and carry no agent phase; action tokens also receive a learned per-step embedding shared by all agents. The action-to-video bias adds the chunk mean of agent $i$'s encoded action tokens to the future tokens of its views, which carry agent $i$'s phase, so each agent's actions are tied to its own views. When roles are defined, the role embedding is added to agent $i$'s view, state, and action tokens and marks them with the agent's function. Agents that share a role receive the same embedding, so their state tokens are told apart only by content (each arm's own normalized proprioception). Each decoded action chunk returns to its agent's controller by index, and relative actions are applied from that agent's own state.

\subsection{\rev{Warm-up and team training}}
\label{app:training}

Training connects single-robot adaptation to joint prediction from synchronized team demonstrations (Table~\ref{alg:training}). Both stages use the world-action objective. Team training starts from the \warmup weights with a fresh optimizer and learning-rate schedule; Table~\ref{tab:stage-config} specifies the inputs and mechanisms active in each stage.

\begin{table}[ht!]
\caption{\rev{Training procedure. Both stages optimize the joint world-action objective.}}
\label{alg:training}
\centering
\small
\begin{tabular}{@{}r >{\raggedright\arraybackslash}p{6.6cm}@{}}
\toprule
 & \textbf{Input:} demonstrations; single-robot pretrained backbone.\\
\midrule
1 & \textbf{\warmup.} Sample one robot's observations, states, and actions; optimize $\mathcal L_{\mathrm{WA}}$.\\
2 & Initialize team training from the \warmup weights.\\
3 & \textbf{\coordstage.} Sample synchronized team data; permute complete agent groups and reassign distinct identity vertices.\\
4 & Optimize $\mathcal L_{\mathrm{WA}}$ jointly over all private-view futures and action tokens.\\
\bottomrule
\end{tabular}
\end{table}

\begin{table*}[ht!]
\caption{\rev{Training configurations. Warm-up adapts the world-action model to individual robots; team training jointly supervises all agents with cross-agent attention, roles, and identities enabled. Both stages use the same world-action loss.}}
\label{tab:stage-config}
\centering
\scriptsize
\setlength{\tabcolsep}{2.4pt}
\begin{tabularx}{\textwidth}{
 >{\raggedright\arraybackslash}p{1.9cm}
 YYYYY}
\toprule
Stage & Parent & Agents per sample & Cross-agent attention & Roles / identities & World-action loss \\
\midrule
\warmup & pretrained backbone & one & off & off & yes \\
\coord & \warmup & all $N$ & full, bidirectional & on & yes \\
\bottomrule
\end{tabularx}
\end{table*}

For \robofactory, the default \warmup budget is 6000 updates. The real-robot tasks use the same world-action objective with their own update budgets.

\subsection{Flow schedules and masked objectives}
\label{app:flow-objective}

Let $\mathbf{x}^{v}_0$ denote clean private-view latents and $\mathbf{x}^{u}_0$ the normalized joint action chunks.  For modality $m\in\{v,u\}$, its scheduler defines
\begin{equation}
 \mathbf{x}^{m}_\tau
 =\alpha^m_\tau\mathbf{x}^{m}_0+
  \beta^m_\tau\boldsymbol{\epsilon}^{m},
 \qquad
 \mathbf{v}^{m,\star}_\tau
 =\dot{\alpha}^m_\tau\mathbf{x}^{m}_0+
  \dot{\beta}^m_\tau\boldsymbol{\epsilon}^{m}.
 \label{eq:forward-process}
\end{equation}
For validity mask $M^m_j$ and per-element weight $w^m_j$, the masked flow-matching loss is
\begin{equation}
 \mathcal{L}_{m}
 =
 \frac{
  \sum_{i=1}^{N}\sum_{j\in\mathcal{J}_i} M^m_j w^m_j
  \left\|\mathbf{v}^{m}_{\theta,j}
            -\mathbf{v}^{m,\star}_{j}\right\|_2^2
 }{
  \sum_{i=1}^{N}\sum_{j\in\mathcal{J}_i} M^m_j w^m_j+\epsilon
 },
 \label{eq:masked-flow-loss}
\end{equation}
where $\mathcal{J}_i$ indexes the elements of agent~$i$ in the mini-batch.
Equation~\ref{eq:world-action-loss} uses $\mathcal{L}_{\mathrm{FM}}(\mathbf O;\mathbf M_v)=\mathcal{L}_v$ and $\mathcal{L}_{\mathrm{FM}}(\mathbf U;\mathbf M_u)=\mathcal{L}_u$.  The video mask excludes observed, padded, and read-only shared global camera elements; the action mask excludes padding and unavailable controller dimensions. Both losses use a single normalization over the valid elements of all agents in the mini-batch, with $w^m_j=1$ in all our experiments. Every agent has the same number of valid action elements (the same 24-step chunk and controller width: 17 dimensions per arm in simulation and 16 logical dimensions per robot on the real robots), so $\mathcal{L}_u$ equals the average over agents of the per-agent masked means.  Video and action noise levels are sampled with modality-specific schedulers.

\section{Evaluation details}
\label{app:evaluation}

\subsection{\rev{Simulation success and uncertainty}}

A crash, timeout, or missing episode counts as a failure. For $k$ successes in $n$ trials, we report the success rate $\hat p=k/n$ in percent with its binomial standard error
\begin{equation}
 \mathrm{s.e.}(\hat p)=\sqrt{\frac{\hat p(1-\hat p)}{n}} .
\end{equation}
Task-group scores are unweighted means of task rates. For equal-sized task sets, the reported uncertainty of the mean is propagated from the task-wise binomial standard errors.

\subsection{\rev{Video-prediction accuracy}}
\label{app:video-accuracy}

We measure how closely the generated private views match the frames the team actually produces. On Lift Barrier, Pass Shoe, Camera Alignment, and Long Pipeline Delivery, we roll out the paper checkpoints on the first 25 evaluation seeds and execute all 24 predicted steps before replanning, so the executed frame at each of $t+3,\dots,t+24$ results from the actions sampled together with the prediction. Frames are compared in the model's input geometry. We compare the prediction with two baselines: the last observed frame, and its VAE reconstruction, which carries the same compression blur as the prediction. Separately, reconstructing the executed target frames at $+12$ and $+24$ with the VAE yields 30.5--34.0\,dB, providing a reference for compression-related error. Most of each view is static: on average 1--6\% of the pixels change within 1.2\,s. Copying the last frame therefore scores higher on full-frame PSNR at most horizons. On frames where at least 5\% of the pixels change (a per-channel difference above 25/255), the prediction is closer to the executed frame by 4--7\,dB at every horizon. The timing column reports, among those frames, how often the executed frame closest to the prediction for $t+h$ is the one at $t+h$ (out of $t,t+3,\dots,t+24$). Camera Alignment's lower timing scores at long horizons come from the hold after the task is complete, where the model predicts a still pose while small corrections keep moving the image; during the task, the score at $+24$ is 62\%.

\begin{table}[ht!]
\caption{\rev{Video-prediction accuracy over 25 seeds per task. PSNR (dB) of the predicted private view at $t+h$ (20\,Hz control steps; $+24$ is 1.2\,s) against the executed frame, for the prediction (Pred.), the last observed frame (Copy), and its VAE reconstruction (Copy-VAE). ``Moving frames'' restricts to frames where at least 5\% of the pixels change. Timing: \% of moving frames whose closest executed frame is at $t+h$.}}
\label{tab:video-accuracy}
\centering
\scriptsize
\setlength{\tabcolsep}{2.5pt}
\begin{tabular}{@{}llcccccc@{}}
\toprule
& & \multicolumn{3}{c}{All frames} & \multicolumn{2}{c}{Moving frames} & \\
\cmidrule(lr){3-5}\cmidrule(lr){6-7}
Task & $h$ & Pred. & Copy & Copy-VAE & Pred. & Copy & Timing\\
\midrule
\multirow{3}{*}{Lift Barrier} & $+6$ & 31.3 & 35.9 & 31.2 & 27.9 & 23.5 & 84\\
 & $+12$ & 30.3 & 34.1 & 30.5 & 26.6 & 20.6 & 69\\
 & $+24$ & 29.7 & 32.2 & 29.6 & 26.5 & 19.5 & 72\\
\addlinespace[2pt]
\multirow{3}{*}{Pass Shoe} & $+6$ & 29.6 & 29.8 & 27.3 & 27.8 & 21.5 & 83\\
 & $+12$ & 28.5 & 27.4 & 25.9 & 27.0 & 21.5 & 75\\
 & $+24$ & 27.2 & 25.1 & 24.3 & 26.2 & 21.3 & 69\\
\addlinespace[2pt]
\multirow{3}{*}{Camera Alignment} & $+6$ & 28.1 & 31.3 & 28.5 & 25.9 & 20.8 & 47\\
 & $+12$ & 26.6 & 27.7 & 26.3 & 24.7 & 19.2 & 20\\
 & $+24$ & 24.4 & 23.4 & 23.1 & 23.6 & 19.2 & 15\\
\addlinespace[2pt]
\multirow{3}{*}{Long Pipeline Delivery} & $+6$ & 31.0 & 34.6 & 30.7 & 27.7 & 22.6 & 68\\
 & $+12$ & 29.8 & 31.7 & 29.5 & 27.2 & 22.3 & 65\\
 & $+24$ & 28.8 & 29.8 & 28.2 & 26.6 & 22.5 & 58\\
\bottomrule
\end{tabular}
\end{table}

\section{\rev{Ablation configurations}}
\label{app:ablations}

The ablation results appear in Section~\ref{sec:ablations} and Table~\ref{tab:ablations}. This section specifies the interventions and evaluation quantities.

\paragraph{Team-stage future-video supervision.}
The \emph{No team-stage future-video supervision} variant sets the future-video loss weight to zero during team training. Everything else is unchanged: Warm-up initialization, demonstrations, training budget, each robot's action tokens, the future-view tokens, the attention masks, and the denoising of all tokens at inference. This control therefore matches \method{} in tokens and test-time compute and differs only in whether the future views are supervised in team training. Both models start from the same Warm-up weights, trained with the future-video loss one robot at a time, so the comparison isolates supervision of the team's joint future views.

\paragraph{Test-time video generation.}
\emph{Frozen future} keeps every future-view token at its initial noise level for all denoising steps, so the token layout, attention masks, and random draws match the full model while no future view is formed. \emph{No future tokens} removes the future-view tokens from the sequence and denoises the actions from the observed views, states, and action tokens alone. Every training sample contains future-view tokens, so this second control also tests an input layout the model never saw. On the real robots, the laundry-bag model takes 0.320\,s per inference with the full model, 0.215\,s without future tokens, and 0.328\,s with a frozen future on one H100.

\begin{table}[ht!]
\caption{\rev{Test-time video generation. Success rates (\%) with binomial standard error over 100 evaluation episodes per task. Differences from the full model are tested with an exact two-sided sign test on paired seeds; no frozen-future change is significant at $p<0.05$ (Place Food 78 vs.\ 70, $p=0.17$). The no-future-tokens control is not available in the Place Food serving stack.}}
\label{tab:test-time-video}
\centering
\footnotesize
\setlength{\tabcolsep}{3pt}
\begin{tabular}{@{}lccc@{}}
\toprule
Task & Full model & Frozen future & No future tokens\\
\midrule
Lift Barrier & \SimMawamLiftPct & \SimTTFrozenLift & \SimTTNoFutLift\\
Pass Shoe & \SimMawamPassShoePct & \SimTTFrozenPassShoe & \SimTTNoFutPassShoe\\
Place Food & \SimMawamPlaceFoodPct & \SimTTFrozenPlaceFood & \SimTTNoFutPlaceFood\\
Two Robots Stack Cube & \SimMawamTwoStackPct & \SimTTFrozenTwoStack & \SimTTNoFutTwoStack\\
Camera Alignment & \SimMawamCameraAlignPct & \SimTTFrozenCameraAlign & \SimTTNoFutCameraAlign\\
Three Robots Stack Cube & \SimMawamThreeStackPct & \SimTTFrozenThreeStack & \SimTTNoFutThreeStack\\
Take Photo & \SimMawamTakePhotoPct & \SimTTFrozenTakePhoto & \SimTTNoFutTakePhoto\\
Long Pipeline Delivery & \SimMawamPipelinePct & \SimTTFrozenPipeline & \SimTTNoFutPipeline\\
\bottomrule
\end{tabular}
\end{table}

\paragraph{Information access.}
Independent agents use shared weights and retain each robot's private views plus the read-only shared global camera, with no cross-agent private-token attention. The \emph{No shared global camera} variant removes the shared global camera during \rev{team training} and evaluation. These interventions address cross-agent token exchange and shared visual context, respectively.

\paragraph{Centralized team.}
The \emph{Centralized team} baseline uses the same backbone, demonstrations, and team-stage training budget as \method, but concatenates all robots' views into one visual context and all arm actions into one vector, without identity codes or role embeddings. Each arm keeps a fixed position in training and evaluation: the same camera slots and the same coordinates in the state and action vectors. In Place Food, Pass Shoe, and Long Pipeline Delivery, each arm performs the same sub-task in every demonstration, so this position also determines its role. To adapt the pretrained single-robot interface, we first warm up the new state/action encoders and action decoder at ten times the base learning rate while updating the backbone at half rate. It proceeds in 2000-update increments until every arm's action loss has settled near the first arm's.
On Pass Shoe and Place Food, 4000 updates sufficed: every arm's action loss was within 0.01 of the first arm's. Long Pipeline Delivery adapted more slowly: at 6000 updates its worst arm's action loss was still 0.34 against 0.03 for the first arm. Team training from this checkpoint yielded \SimCentralPipelineSixK{}. To examine whether insufficient interface adaptation accounted for this result, we extended interface warm-up by another 2000 updates, keeping the optimization settings unchanged.
After 8000 updates, all four arms' late-stage action losses were within 0.07 of the first arm's, and the team stage trained normally: the arms' action losses stayed close together (worst arm 0.08 at update 1000) and the joint commands were smooth (0.02--0.07\,rad per step at the first checkpoint). Table~\ref{tab:ablations} reports this run; the evaluation set is the same as for every other row.

\paragraph{Agent identity.}
Absolute slots use learned ordered slot identities with a fixed assignment. Fixed simplex retains the simplex coordinates with a fixed assignment. The full model reassigns distinct simplex vertices to complete agent groups. The full model and Fixed simplex therefore share the code and differ in the assignment, and Fixed simplex and Absolute slots share the fixed assignment and differ in the code. Success rates for all three variants on Lift Barrier, Place Food, Camera Alignment, and the four-arm Take Photo task are reported in Table~\ref{tab:ablations}.

\paragraph{Task roles.}
The no-roles variant assigns the same role label to every agent. Like the full model, it randomizes identity assignment and shuffles complete agent groups in training, so no identity label or sequence position stays tied to an arm; the centralized team instead keeps every arm in one position. Camera Alignment's shared-holder variant uses roles $[0,0,1]$ instead of the main configuration's $[0,1,2]$, grouping the two camera holders by function. Role-only swaps exchange asymmetric role labels at evaluation while fixing the observations and controller destinations. They test sensitivity to the supplied assignment under the original task's success criteria and give \SimRoleSwapPlaceFood\% on Place Food and \SimRoleSwapCameraAlign\% on Camera Alignment.

\paragraph{Response to a delayed teammate.}
The hold uses the same 100 Long Pipeline Delivery evaluation seeds as Table~\ref{tab:sim-results}. When the giver releases the shoe at the second handover, the evaluation replaces its commands with its current joint positions and an open gripper for 16 steps; the policy then resumes control. The receiver's start is the first step after the release at which its joints move faster than 0.002\,rad per step for at least three consecutive steps. One MAWAM seed, its only failure under the hold, never reaches the second handover, so it has no release, hold, or start delay; MAWAM's 99 delays and 99 successes therefore come from the same episodes, and independent agents have a delay on all 100 seeds. MAWAM's receiver also grasps later (median $+9$ steps; independent agents $+1.5$), no episode has contact between the two arms, and success changes from 95 to 99 for MAWAM and from 39 to 44 for independent agents. Excluding the 18 episodes in which MAWAM's giver grasps the shoe again after the hold, the median start delay is 10 steps. Of the 99 paired seeds, MAWAM's delay is longer on 86, shorter on 10, and equal on 3; an exact two-sided sign test excluding ties ($n=96$) gives $p=3.2\times10^{-16}$.

\paragraph{Warm-up.}
The no-Warm-up configuration starts team training directly from the pretrained backbone. The full configuration initializes team training from the \warmup weights.

\section{Real-robot data and evaluation details}
\label{app:real}

\begin{table*}[ht!]
  \caption{\rev{\textbf{Model size and deployment schedule of the real-robot policies.} Trainable params counts the parameters updated during training; frozen pretrained encoders are excluded. Horizon gives predicted and executed action steps per inference at 15\,Hz. Latency is the median model time over 60 inferences after warm-up on one NVIDIA H100 (fp32). \method generates the team action chunks in two joint denoising forwards.}}
  \label{tab:real-efficiency}
  \centering
  \scriptsize
  \setlength{\tabcolsep}{3pt}
  \begin{tabularx}{\textwidth}{
    >{\raggedright\arraybackslash}p{2cm}
    >{\raggedright\arraybackslash}p{3.4cm}
    >{\raggedright\arraybackslash}X
    >{\raggedright\arraybackslash}p{2.4cm}
    >{\centering\arraybackslash}p{1.2cm}}
  \toprule
  Method & Trainable params & Sampling per inference & Horizon (pred.\ $\rightarrow$ exec.) & Latency \\
  \midrule
  CLS-DP & 244M (SigLIP-B/16 frozen)
  & 100-step DDPM: 1 contextualizer forward + 100 UNet forwards per agent, both agents on one GPU
  & $8 \rightarrow 6$ (0.4\,s) & 0.77\,s \\
  GauDP & 131M (NoPoSplat encoder frozen) & 100-step DDPM: 1 NoPoSplat encoder forward + 100 UNet forwards & $8 \rightarrow 6$ (0.4\,s) & 0.48\,s \\
  MA-VLA ($\pi_{0.5}$) & 3.35B (full fine-tuning) & 10-step flow: 1 prefix forward + 10 action-expert forwards & $50 \rightarrow 8$ (0.53\,s) & 0.065\,s \\
  \rowcolor{ourshade}\method (ours) & 2.76B (2.06B DiT + 0.70B state/action encoders and action decoder; Cosmos-Reason1 and VAE frozen) & 2-step flow matching: 2 joint forwards & $24 \rightarrow 8$ (0.53\,s) & 0.32\,s \\
  \bottomrule
\end{tabularx}
\end{table*}

\subsection{Hardware and setup}
\label{app:real-setup}

\paragraph{Robots.}
We use two Galaxea R1 Pro robots per task. Each robot has two 7-DoF arms with two-finger grippers, a 4-DoF torso, and a wheeled base; the base and torso are not commanded by the policy, and the policy sends end-effector poses and gripper commands only. The two robots in any pair face each other.

\paragraph{Initial pose.}
Before every episode both robots are driven to the same fixed pose (Table~\ref{tab:real-home}). The torso leans forward and is then held fixed for the whole episode. Each arm starts with the elbow at about $-93^\circ$ and the wrist rotated $\pm 90^\circ$, which places both grippers about 0.42\,m in front of the torso and 0.25\,m to either side, level with the table.

\begin{table}[ht!]
\caption{Initial joint configuration used for each of the two robots.
Joint angles are given in radians, with degree equivalents in parentheses.
Left and right refer to the arms of each robot; arm joints are listed in J1--J7 order.}
\label{tab:real-home}
\centering
\scriptsize
\setlength{\tabcolsep}{3pt}
\renewcommand{\arraystretch}{1.1}
\begin{tabularx}{\columnwidth}{
  >{\raggedright\arraybackslash}p{2.5cm}
  >{\raggedright\arraybackslash}X
}
\toprule
Group & Joint targets \\
\midrule
Torso (4 joints)
& $0.451,\ -0.750,\ -0.850,\ 0.000$ \newline
  ($25.8^\circ,\ -43.0^\circ,\ -48.7^\circ,\ 0^\circ$) \\

Left arm
& $0,\ 0,\ 0,\ -1.620,\ +\pi/2,\ 0,\ 0$ \newline
  (J4: $-92.8^\circ$; J5: $+90^\circ$) \\

Right arm & $0,\ 0,\ 0,\ -1.620,\ -\pi/2,\ 0,\ 0$  \newline
  (J4: $-92.8^\circ$; J5: $-90^\circ$) \\

Both grippers & Fully open \\
\bottomrule
\end{tabularx}
\end{table}

\paragraph{Cameras.}
Each robot contributes three RGB streams at 15\,Hz: the left image of the stereo head camera (640$\times$360, center-cropped to 480$\times$360) and one Intel RealSense D405 on each wrist (640$\times$480). All six streams are center-cropped to 95\% and resized to 224$\times$320 (height$\times$width) before entering the model, which sees five frames per stream spanning the last 0.8\,s. No shared global camera is used.

\paragraph{Control and serving.}
The policy outputs, per robot, a relative end-effector pose (position and quaternion) for each arm and one opening value per gripper, at 15\,Hz in 24-step chunks. Commands go to the vendor inverse-kinematics and joint tracker on each robot. One synchronized coordinator collects both robots' observations, queries the policy server (one NVIDIA H100) over the network, and dispatches both action chunks together; the compiled model uses 0.32\,s for two joint denoising forwards. The controllers execute the first 8 steps of each 24-step chunk in lockstep, observe, and replan (Table~\ref{tab:real-efficiency}).

\paragraph{Synchronized demonstrations.}
Demonstrations are collected by one operator per robot using a Meta Quest headset and controllers through the vendor's VR interface, and recorded at 15\,Hz with the same three cameras per robot. For umbrella handover, each pair consists of one giver and one receiver. Their role labels are attached to the source trajectories and preserved when complete agent groups are reordered. For laundry-bag collection and bath-towel folding, the role vocabulary and mapping are defined by the collection protocol and fixed before training. Before splitting the data, we store task identities in the pair metadata and exclude records with missing source identifiers or required camera streams.

\subsection{Post-training and deployment}
\label{app:real-deployment}

On the real robots, \method training starts from GR00T~2 and uses the same two-stage recipe and world-action objective as simulation. The single-robot \warmup uses the same three private views and 16D logical action definition as team training. Team training receives synchronized team trajectories and supervises the futures of all six private views together with both robots' actions. Both stages use the robot-mounted cameras without introducing a shared global camera.

At inference, one joint sampling call with two denoising forwards generates private-view futures and both robots' action chunks from the synchronized six-view input. Before execution, we jointly validate timestamp skew, ownership and role metadata, validity masks, finite command values, controller ranges, command age, workspace limits, and collision status. A failed check triggers a coordinated stop for both robots.

\subsection{Task success and safety accounting}
\label{app:real-success}

Before every trial, both robots return to the fixed pose in Table~\ref{tab:real-home}. We randomize the initial object pose, garment position, and folding direction within a certain range. Every model is evaluated for 30 trials per task. The task-specific terminal success criteria are:
\begin{itemize}
 \item \textbf{Umbrella handover:} the receiver stably holds the umbrella, and the giver has released it.
 \item \textbf{Laundry-bag collection:} the garment is inside the bag at termination, and neither robot still grasps it.
 \item \textbf{Bath-towel folding:} the towel is folded in the intended direction with its edges and corners aligned, and both robots have released it.
\end{itemize}
We report the success rate with its standard error over 30 trials per task.

\subsection{\rev{Model size and inference schedule}}
\label{app:real-cost}
Table~\ref{tab:real-efficiency} specifies how each deployed policy converts one inference call into an executable action horizon. \method's model time covers two joint denoising forwards; the 8 executed steps correspond to 0.53\,s of motion at 15\,Hz. Network transport and physical execution are outside this model-time measurement.

In simulation, one joint sample serves the whole team, and its cost follows the number of camera views (Table~\ref{tab:sim-latency}). Doubling the team from two to four arms adds about a third to the time per sample (1.40\,s to 1.85\,s), and removing the shared global camera from a two-arm team brings it to 0.76\,s. Fewer denoising steps shorten it further: Camera Alignment uses two steps and takes 0.81\,s.

\begin{table}[ht!]
\caption{\rev{\textbf{Simulation inference latency.} Median model time per joint sample on one NVIDIA L40 (bf16, four denoising steps, no compilation). View slots count the model's camera inputs; two-arm models with the shared global camera pad to four slots. Real-robot timing (Table~\ref{tab:real-efficiency}) uses an H100, two steps, and a compiled DiT.}}
\label{tab:sim-latency}
\centering
\scriptsize
\setlength{\tabcolsep}{3pt}
\begin{tabularx}{\columnwidth}{ccc>{\raggedright\arraybackslash}Xc}
\toprule
Arms & \shortstack{Shared global\\camera} & \shortstack{View\\slots} & Tasks & \shortstack{Latency\\(s)} \\
\midrule
2 & -- & 2 & Lift Barrier, Place Food, Two Robots Stack Cube & 0.76 \\
2 & \abon & 4 & Lift Barrier, Place Food, Pass Shoe, Two Robots Stack Cube & 1.40 \\
3 & \abon & 4 & Three Robots Stack Cube & 1.44 \\
4 & \abon & 5 & Take Photo, Long Pipeline Delivery & 1.85 \\
\bottomrule
\end{tabularx}
\end{table}

\clearpage
{
    \emergencystretch=1em
    \setcitestyle{numbers}
    \renewcommand{\natexlab}[1]{#1}  
    \bibliographystyle{ieeenat_fullname}
    \bibliography{references}
}

\end{document}